\documentclass{article} 
\usepackage{iclr2025_conference,times}
\usepackage{algorithm}
\usepackage{algpseudocode}
\usepackage{makecell}

\usepackage{amsmath,amsfonts,bm}

\def\eqref#1{equation~\ref{#1}}

\def\1{\bm{1}}

\DeclareMathAlphabet{\mathsfit}{\encodingdefault}{\sfdefault}{m}{sl}
\SetMathAlphabet{\mathsfit}{bold}{\encodingdefault}{\sfdefault}{bx}{n}

\definecolor{nicepurple}{RGB}{148,0,211}

\usepackage{url}
\usepackage{graphicx}

\usepackage[colorlinks = true,
            linkcolor = black,
            urlcolor  = blue,
            citecolor = teal,
            anchorcolor = blue]{hyperref}

\usepackage{enumitem}

\title{Generating Freeform Endoskeletal Robots}

\author{Muhan Li,\, 
Lingji Kong,\, Sam Kriegman \\
Northwestern University
}

\iclrfinalcopy 
\begin{document}

\maketitle

\thispagestyle{empty}

\begin{figure}[h]
    \centering

    \vspace{-16pt}
    \includegraphics[width=\textwidth]{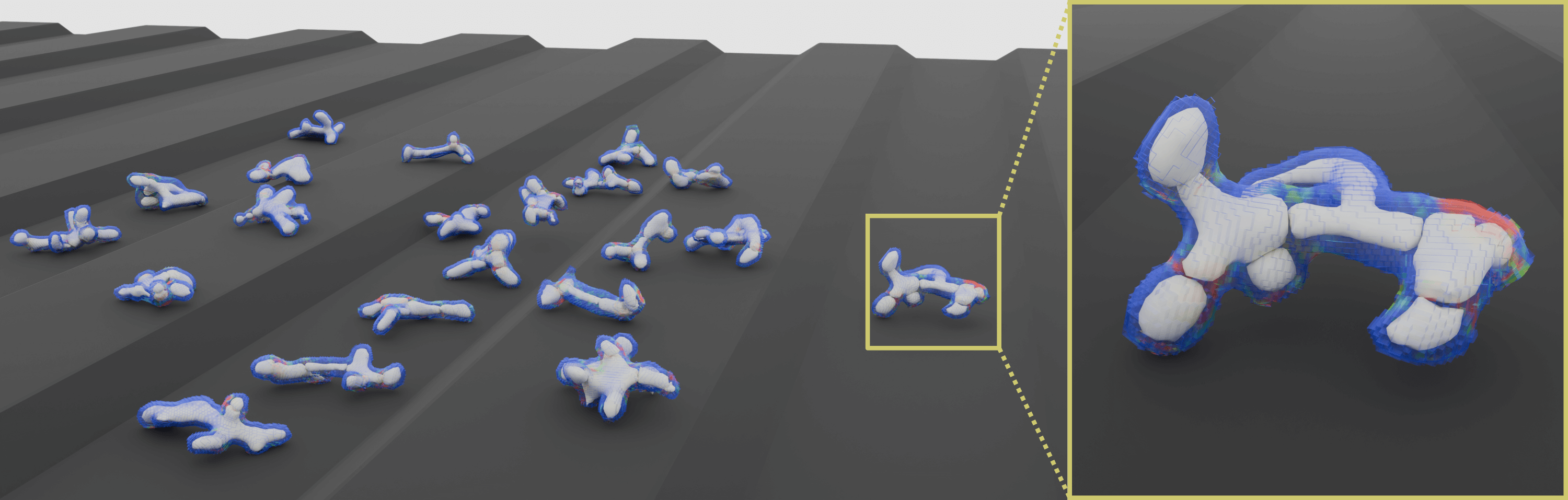}

    \vspace{-8pt}
    
    \caption{An 
    evolving population 
    of endoskeletal soft robots
    were encoded by a
    low dimensional latent design genome 
    with minimal morphological assumptions
    and optimized for 
    locomotion across 
    complex terrains
    in multiphysics simulation
    using a shared universal controller that was simultaneously 
    learned alongside morphological design.
    Videos and code at \href{https://endoskeletal.github.io}{\color{blue}https://endoskeletal.github.io}.
}
    \label{fig:teaser}
\end{figure}

\begin{abstract}

The automatic design of embodied agents (e.g.~robots) 
has existed for 31 years and
is experiencing 
a renaissance of interest in the literature.
To date however,
the field has remained narrowly focused on
two kinds of anatomically simple~robots: 
(1)~%
fully rigid, jointed bodies;
and 
(2)~%
fully soft, jointless bodies.
Here we 
bridge these two extremes with
the open ended creation of 
terrestrial 
endoskeletal robots:
deformable soft bodies
that leverage
jointed internal skeletons 
to move efficiently
across land.
Simultaneous de novo generation 
of external and internal structures
is achieved by 
(i)~%
modeling 
3D endoskeletal body plans 
as 
integrated 
collections of 
elastic and rigid cells
that directly attach
to form
soft tissues
anchored to
compound rigid bodies;
(ii)~%
encoding these
discrete mechanical subsystems 
into a 
continuous yet
coherent
latent embedding;
(iii)~%
optimizing the sensorimotor coordination of 
each decoded design using 
model-free
reinforcement learning;
and 
(iv)~%
navigating this
smooth yet
highly non-convex
latent manifold using 
evolutionary strategies.
This 
yields an 
endless stream of 
novel species of 
``higher robots'' 
that, 
like all higher animals,
harness 
the 
mechanical advantages 
of both 
elastic tissues
and 
skeletal levers
for terrestrial travel.
It also provides 
a plug-and-play 
experimental 
platform
for benchmarking
evolutionary design 
and
representation learning 
algorithms in complex hierarchical embodied systems.
%
%

\end{abstract}

\begin{figure}[t]
    \centering
    \includegraphics[width=\textwidth]{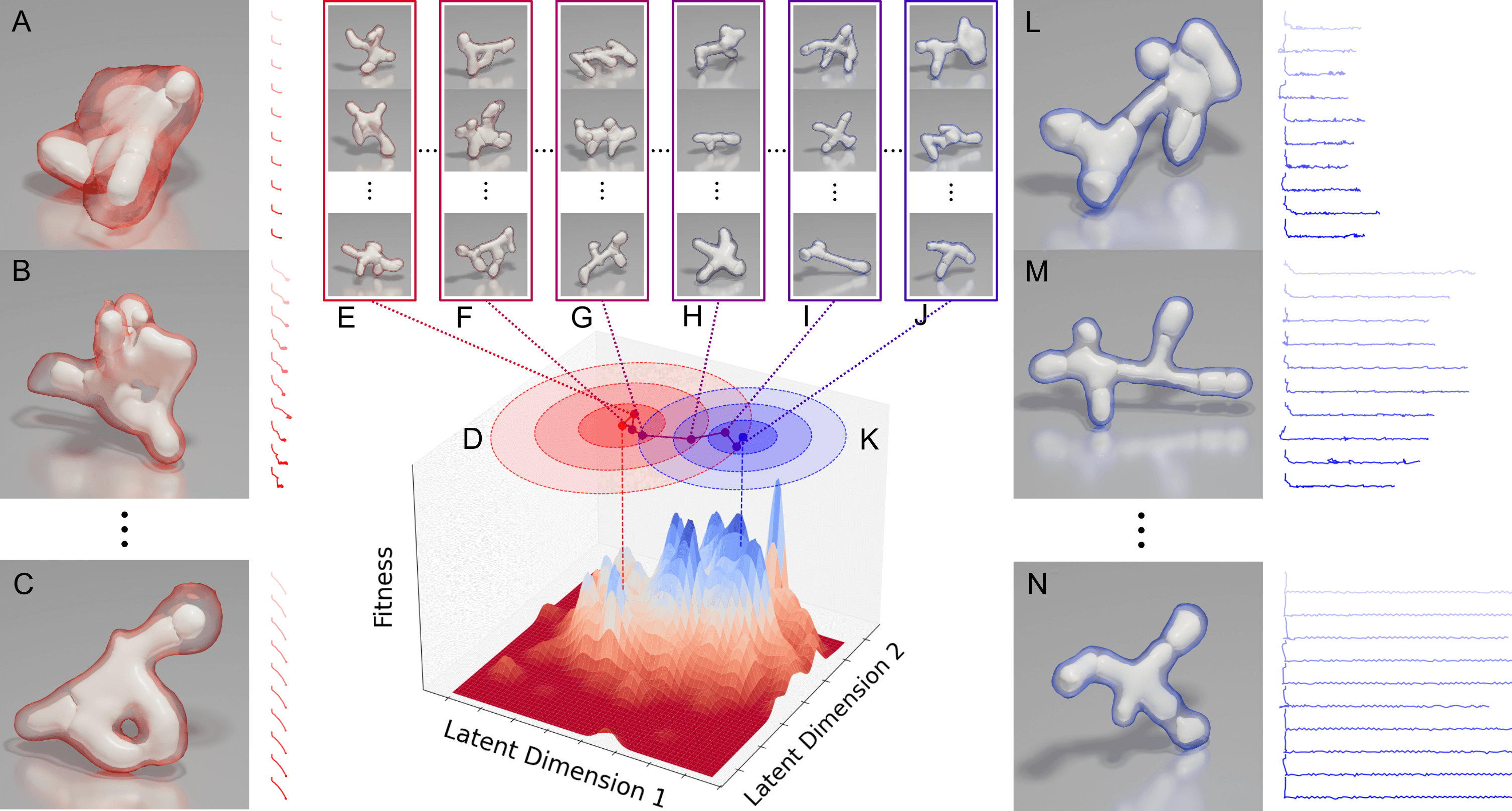}
    \vspace{-14pt}
    \caption{\textbf{Designing endoskeletal robots.} 
    A population of 64 robot designs (\textbf{A-C}) was sampled from a multivariate normal distribution (\textbf{D}) centered about an initially random point in latent space.
    The behavior each design in the population
    was optimized
    in physics simulation
    for 20 epochs of reinforcement learning 
    under a shared universal controller.
    The ten red behavioral traces 
    to the right of A-C project the forward locomotion
    of these randomly generated designs
    (from 3D to 2D)
    into the right hand side of the page, at every other epoch of learning.
    The fitness of a design was based on the best performance it achieved during reinforcement learning.
    After learning, the mean and covariance matrix of the design distribution 
    were then adapted by evolutionary strategies
    and a subsequent generation of 64 new designs was sampled from the shifted distribution (\textbf{E}).
    This process was repeated for dozens of generations (\textbf{F-J}) 
    yielding 
    an evolved design distribution
    (\textbf{K})
    that encodes
    a population of designs (\textbf{L-N}) 
    with much higher fitness (blue behavioral traces).
    \vspace{-6pt}
}
    \label{fig:evolution}
\end{figure}

\begin{figure}[t]
    \centering
    \includegraphics[width=\textwidth]{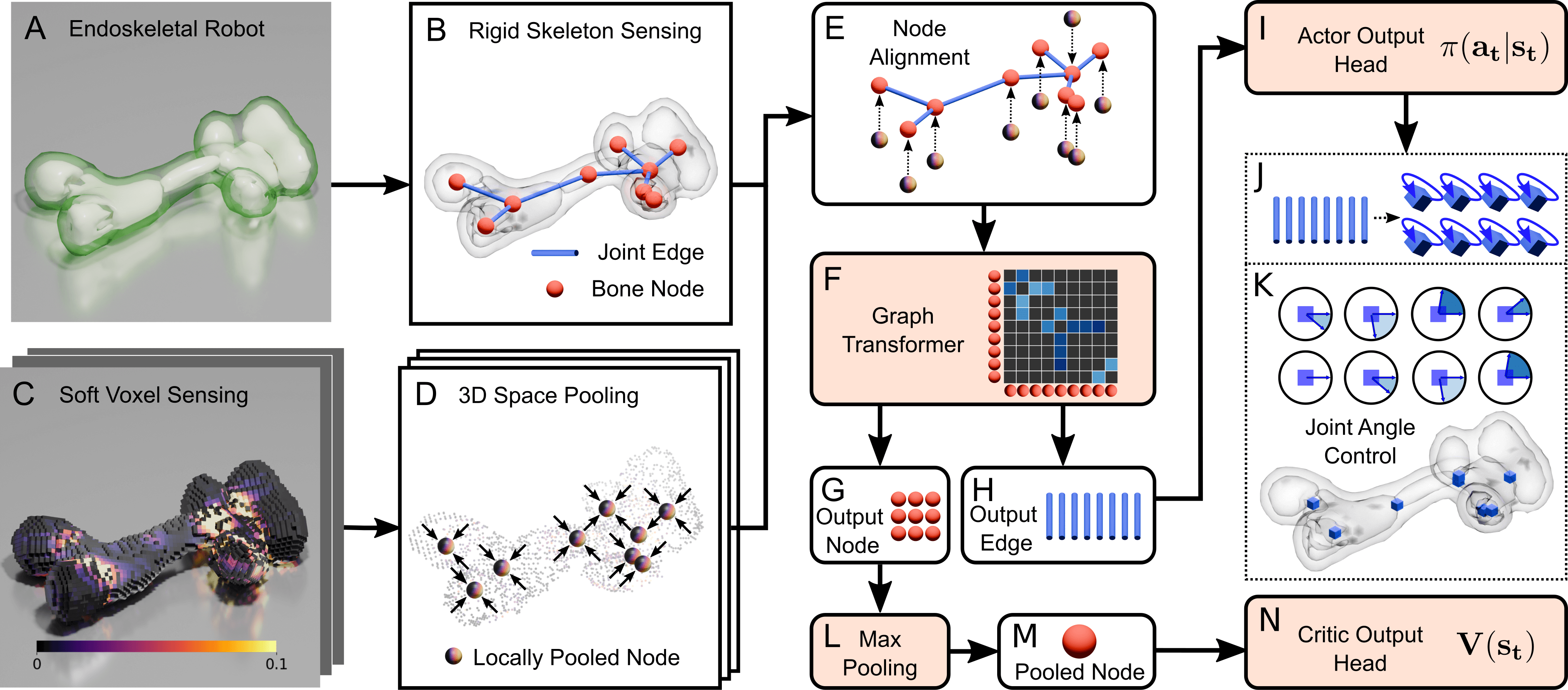}

    \vspace{-2pt}
    
    \caption{\textbf{A universal controller for freeform endoskeletal robots.}
    The control of an endoskeletal robot (\textbf{A})
    utilizes 
    a graph data structure
    with edges that map
    bone connections within its skeleton,
    edge features that 
    track the position, orientation and angle of each joint,
    and
    node features
    that
    track the position, orientation and movement 
    of each bone 
    (\textbf{B}).
    The controller also tracks 
    proprioceptive and mechanosensory input from the position, velocity and strain of the robot's soft tissues (\textbf{C}),
    which is locally pooled 
    into the center of mass of each bone (\textbf{D}) thereby 
    transforming the soft tissue's sensory input from voxel space to a graph that aligns with the skeletal sensory graph (\textbf{E}).
    The combined rigid+soft 
    sensory graph is then fed as input to a graph transformer (\textbf{F}).
    The graph transformer distils sensory signatures across the graph 
    into updated node features (\textbf{G}), 
    and updated edge features (\textbf{H}). 
    The Actor (\textbf{I}) takes 
    the updated edge features as input and 
    outputs motor commands (\textbf{J}):
    the target rotation angle for each joint (\textbf{K}).
    The updated node features are pooled by
    globally by a channel-wise
    maximum across the node dimension
    (\textbf{L})
    to retrieve a graph-level output (\textbf{M}), 
    which the Critic (\textbf{N}) uses to predict a value function based on the robot's current state information.
    \vspace{-16pt}
}
    \label{fig:RL}
\end{figure}

\section{Introduction}
\label{sec:intro}

The manual synthesis of 
a single robot 
for a single task 
requires 
years if not decades of 
labor-intensive R\&D
from teams of engineers
who
rely
heavily on expert knowledge, intuition, and hard-earned experience.
The automatic synthesis of robots
may
reveal
designs 
that are different from 
or beyond what human engineers 
were previously capable of imagining
and
would be of great use if robots are to be designed and deployed for many disparate tasks in society.
However, automating the 
creation and improvement 
of a robot's design 
is notoriously difficult
due to 
the 
hierarchical and combinatorial 
nature of the optimization problem:
the objective and variables of 
the robot's 
controller optimization 
depend on
the optimizer of the robot's 
discrete 
mechanical structure
(e.g.~its distribution of motors and limbs).
Considerable effort has been and is being expended to 
ameliorate 
\citep{zhao2020robogrammar,yuan2022transformact,strgar2024evolution}
or 
obviate 
\citep{iii2021taskagnostic,gupta2022metamorph,matthews2023efficient} 
these technical challenges.
Nevertheless,
automatically designed robots
have yet to evolve beyond
primitive machines
with 
exceedingly simple body plans, 
which
can be neatly divided into
two opposing groups:

\vspace{-4pt}
\begin{enumerate}


    \item fully rigid, jointed bodies \citep{sims1994competition,lipson2000automatic,auerbach2014environmental,zhao2020robogrammar,iii2021taskagnostic,gupta2021embodied,yuan2022transformact}; and

    \item fully soft, jointless bodies \citep{hiller2012automatic,cheney2018scalable,kriegman2020xenobots,matthews2023efficient,wang2023diffusebot,li2024reinforcement,strgar2024evolution}.

\end{enumerate}
\vspace{-4pt}

\begin{figure}[t]
    \centering
    \includegraphics[width=\textwidth]{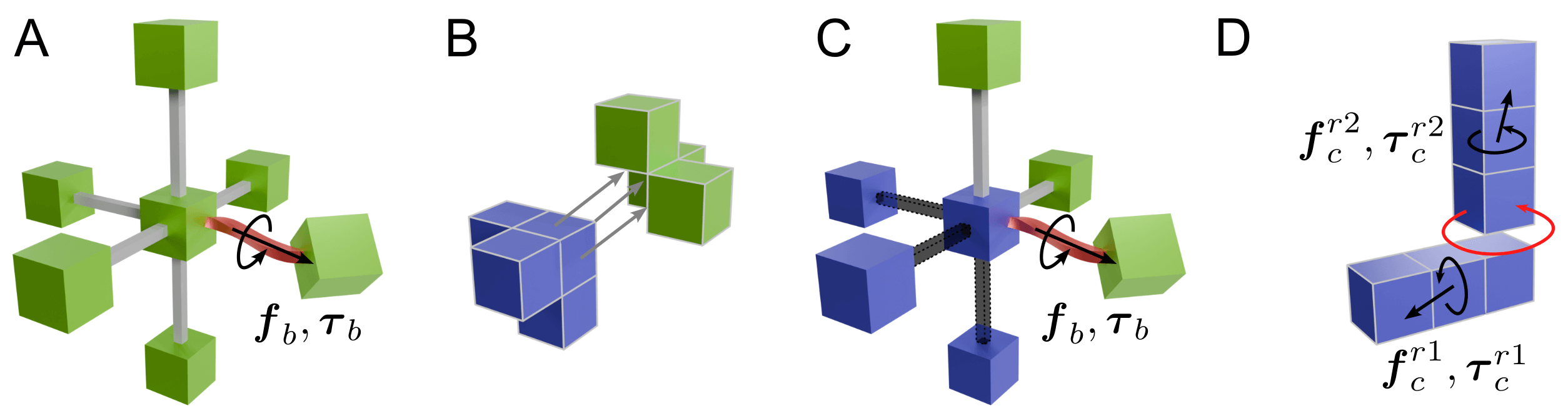}
    \vspace{-16pt}
    \caption{\textbf{Simulating endoskeletal robots.}
    Soft tissues (green masses) were modeled by a grid of Euler–Bernoulli beams (\textbf{A})
    that may twist and stretch
    and directly attach to bones (blue masses; \textbf{B} and \textbf{C}) that follow rigid bodied dynamics with joint constraints
    (\textbf{D}).
    More details about the simulator
    can be found in 
    Sect.~\ref{sec:sim}
    and 
    Appx.~\ref{appx:sim}.
    \vspace{-8pt}
}
    \label{fig:sim}
\end{figure}

Here we introduce  
the de novo optimization
of
externally soft 
yet
internally rigid, jointed body plans---%
endoskeletal robots (Fig.~\ref{fig:teaser})---%
with minimal assumptions about the 
robots' external and internal structures (Fig.~\ref{fig:evolution}).
Endoskeletal robots contain 
bones and joints---the 
lever systems animals use
to their mechanical advantage
as they gallop across uneven surfaces,
climb up trees, and 
swing from branch to branch.
At the same time,
endoskeletal robots contain and are encased by 
soft deformable tissues---the springs 
animals use to
conform to uneven surfaces, 
absorb shocks, 
store and release elastic energy,
catapult themselves into the air,
and capture
complex somatosensory signals 
about their environment.
In order to achieve this 
confluence of 
soft and rigid body dynamics 
within a
single robot---%
without presupposing 
the
topology or
the geometry 
of the robot's 
bones and tissues---%
a novel 
multi-physics 
voxel-based simulator (Fig.~\ref{fig:sim})
was created and is introduced here 
alongside 
a latent endoskeletal embedding
and companion
generator 
that 
sythensizes 
elastic and rigid voxel cells 
into 
freeform tissues and bones,
and connects the bones by joints
within
a functional 
whole.

Several existing simulators support 
two-way coupling between  
rigid and deformable bodies with joint constraints
\citep{shinar2008two,kim2011fast,li2020soft}.
%
However they are not 
naturally amenable to
structural optimization 
because they do not readily expose building blocks (e.g.~voxels) 
to the optimizer; 
instead, 
they require a 
bespoke volumetric mesh 
to be supplied for each body plan, 
and the mesh
must be carefully designed 
to remain numerically stable.
\cite{jansson2003combining}
embedded rigid bodies within a regular grid of
passive Hookean springs and masses, 
which provides suitable building blocks of freeform structure,
but does not support joint constraints, actuation, 
volume preservation under twisting deformations, 
or the parallelization necessary to
scale design to complex bodies composed of large numbers of building blocks.
Here, in contrast, we 
present a massively-parallel 
simulator that uses 
a more accurate, twistable 
model of elasticity
\citep{hiller2014dynamic}
as the basis of
building blocks 
of soft tissues
which directly attach to
rigid building blocks of bones
and articulated skeletons
with joint constraints 
and actuation 
\citep{baraff2001physically}.

\cite{cheney2013unshackling}
evolved robots composed of
elastic building blocks
they called ``muscle'', ``fat'' and ``bone''.
The muscles rhythmically pulsed, expanding and contracting in volume;
the fat was passive;
and the ``bone''
was equivalent to fat 
in density and mass
but had 
an order-of-magnitude higher 
stiffness (Young's modulus).
They found that any further increases to bone stiffness
prevented 
the pulsating muscle blocks from sufficiently deforming the robot's jointless body
and thus inhibited locomotion.
In reality, animal bones 
are seven orders-of-magnitude stiffer than animal fat \citep{guimaraes2020stiffness}, 
and are pulled by tendons and muscles about joints to rapid generate large yet precise dynamic movements 
of load bearing limbs.
One could simply add joints to this elastic model; 
however, 
accurate simulations of
elastic elements
with higher stiffness
(and thus higher resonance frequencies)
approaching that of animal bones
quickly become prohibitively expensive
as ever smaller time steps are required
to avoid numerical instability.
Moreover,
the fully ``soft'' (entirely non-rigid) and jointless
robots evolved by 
\cite{cheney2013unshackling} also
lacked sensors,
were composed of no more than a few hundred elastic blocks,
and could only move along a perfectly flat surface plane.
Here, in contrast,
we introduce a new approach
that combines
hundreds of thousands of 
both 
elastically deformable and rigid 
building blocks
into
multi-physics bodies 
that use
joints and sensors
to navigate complex terrestrial environments.

\begin{figure}[t]
    \centering
    \includegraphics[width=\textwidth]{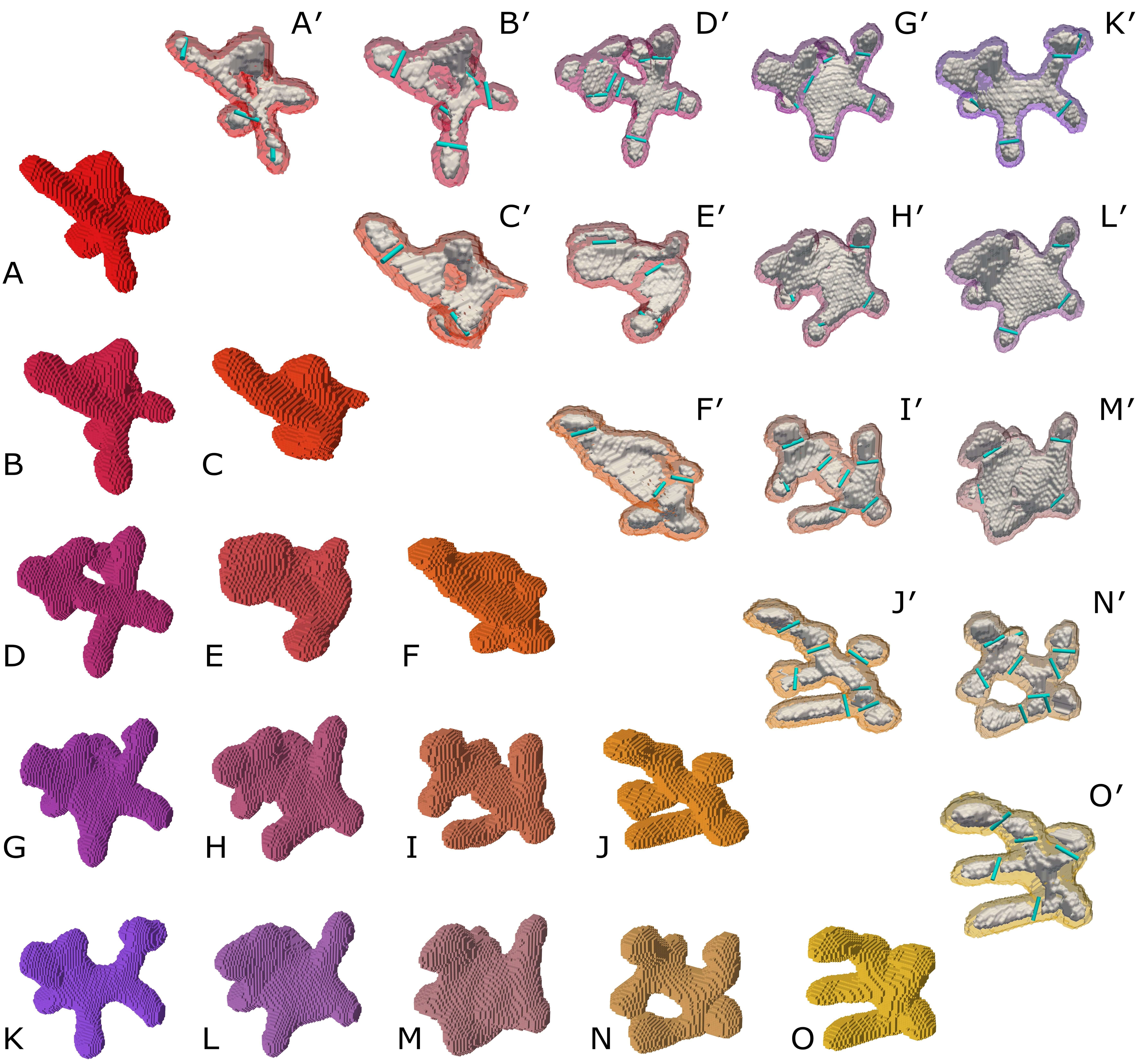}
    \vspace{-12pt}
    
    \caption{\textbf{Interpolating between three points in endoskeletal latent space.}
    Designs sampled from a 2D slice of the
    learned latent embedding (\textbf{A-O})
    and their internal jointed skeletons
    (\textbf{A$^\prime$-O$^\prime$}),
    where A$^\prime$ reveals the skeleton of A and cyan cylinders show the location of hinge joints between bones in each skeleton.
    The three corner designs (A,K,O) were drawn from arbitrarily selected latent coordinates 
    and the 12 designs between them sit with equal spacing along a plane of linear interpolation. 
    The visible smoothness,
    mechanical coherence,
    and
    geometric and topological expressiveness of the voxel-based latent space facilitated the co-design of morphology and control. 
    \vspace{-8pt}
}
    \label{fig:latent}
\end{figure}

Following \cite{hiller2010evolving},
Compositional Pattern Producing Networks (CPPNs; \cite{stanley2007compositional})
have been used by
many others 
\citep{cheney2013unshackling,auerbach2014environmental,cheney2018scalable,kriegman2021kinematic,wang2023softzoo,cochevelou2023differentiable}
to encode robot designs.
CPPNs are a class of 
feedforward neural networks with spatial inputs (e.g.~cartesian coordinates) 
and hence spatially correlated outputs, 
which can be used to determine the absence or presence of a cell at each queried spatial coordinate.
CPPNs
thus provide
a simple way to 
``paint'' material 
into a workspace;
however, 
they struggle to 
integrate basic 
structural and functional constraints, 
such as those 
which ensure that
rigid cells belonging to the same bone form a single contiguous object,
that joints are only placed between nearby bones,
and that bones remain inside the body.
Moreover,
one CPPN 
(its entire architecture)
encodes a single design
rather than
a latent manifold of
continuous variables 
that encode a generative model of robot designs.
Here 
we learn one such 
continuous representation 
for 
endoskeletal robots
and demonstrate the learned representation's
structural and functional
coherence, expressivity, smoothness  (Fig.~\ref{fig:latent})
\mbox{and
interpretability 
(Figs.~\ref{fig:latent_correlation}-\ref{fig:gene_editing}).}

Prior work encoded robot designs 
into a continuous latent embedding \citep{hu2023glso};
however, the design space 
\citep{zhao2020robogrammar}
was limited by graph grammars
to simple body plans
containing
a small number of jointed rigid links
with predefined shapes.
These simplifying assumptions 
allowed for
effective 
model predictive control \citep{lowrey2018plan}, 
which requires an explicit and accurate model of the robot's dynamics, 
including its interaction with the environment.
Here, in contrast, we use a
model-free, end-to-end learning approach
to design and control 
externally-soft body plans
with highly 
non-linear
dynamics
and
multi-physics
interactions
that are 
difficult to model
explicitly.

More specifically, we here train a single reinforcement learning system---a ``universal controller'' (Fig.~\ref{fig:RL})---%
across 
an evolving population of freeform endoskeletal robots. 
%
This is achieved by
locally pooling 
\citep{peng2020convolutional}
each robot's arbitrary 
arrangement of
high-resolution (voxel-level) 
sensory channels
to align with its skeletal graph, 
which may then be fed as input to a graph transformer \citep{yun2019graph}.
Graph transformers were previously employed
for universal control of many 
robots with differing physical layouts \citep{gupta2022metamorph};
however,
these robots
were constrained to
bilaterally symmetrical ``stick figures'' composed of jointed 
rigid cylinders \citep{ventrella1994explorations}.
There was no need 
to learn a 3D volumetric representation
of sensor integration
as the stick figures 
were already in a simple graph form with their sensors 
directly aligned 
to the edges of their graph.
We here evolve embodied agents
beyond the rigid stick figures
\citep{yuan2022transformact,gupta2022metamorph}
and boneless ``blobs'' 
\citep{wang2023softzoo,huang2024dittogym}
found in recent work
with the multiphysics simulation,
freeform design
and universal control
of endoskeletal robots.

\section{Methods}
\label{sec:methods}

In this section we describe how
endoskeletal robots were simulated (Sect.~\ref{sec:sim}), 
encoded (Sect.~\ref{sec:ecoding}) 
and optimized (Sect.~\ref{sec:codesign}).
A symbol table is provided on the very last page of the paper, in Table.~\ref{table:symbol_table}.

\subsection{Simulating Endoskeletal Robots}
\label{sec:sim}

We here introduce a novel voxel-based experimental platform that was 
built from the ground up 
to support the development of novel representations and algorithms 
for co-designing the form and function of endoskeletal soft robots. 
Fig.~\ref{fig:sim}
provides a high level overview of the underlying physics simulator, which enables
two-way coupling between freeform soft tissues
and 
freeform rigid bones  
with
joint constraints.
In this voxel-based simulator,
elastic voxels and rigid voxels are used as building blocks of soft tissues
and bones, respectively.
Elastic voxels and rigid voxels are
subject to their own respective
update rules,
which are outlined in Appx.~\ref{appx:sim}.

The most challenging part of building such a simulator was integrating the dynamics of soft and rigid components,
which can be summarized by a single equation for all voxels:
\begin{equation}
    M^v \ddot{X}^v = F_b^v + F_{\text{ext}}^v + \mathbb{I}(v \in \text{bone}) F_c^v ,
\end{equation}
where $v$ is a single voxel, 
$M^v \in \mathbb{R}^{6 \times 6}$ is the generalized mass matrix combining mass and inertia terms, 
$\ddot{X}^v \in \mathbb{R}^{6}$ is the second derivative of generalized position $X^v$ with respect to time $t$, 
$F_b^v \in \mathbb{R}^{6}$ and $F_{\text{ext}}^v \in \mathbb{R}^{6}$ are the generalized force terms including force and torque terms, 
$F_b^v$ is the effect of all elastic beams (Fig.~\ref{fig:sim}A) 
and $F_{\text{ext}}^v$ is the effect of all external forces such as gravity and friction. $F_c$ is the generalized constraint force term which is only applied on rigid voxels using the indicator function $\mathbb{I}(v \in \text{bone})$.
For more details see Appx.~\ref{appx:sim}.

\subsection{Encoding Endoskeletal Morphology}
\label{sec:ecoding}

We began by procedurally generating synthetic training data---examples of valid endoskeletal body plans---in 3D voxel space using multi-star graphs (Fig.~\ref{fig:synth_vs_decode}A), which consist of random number of randomly positioned
star nodes
like this {\Large$\ast$}
with a variable number of arms radiating from their center points.
The radius and length of the bone and the radius of the surrounding 
soft tissue in each
arm were all randomly assigned.
Each arm may connect by a joint
to an arm of an adjacent star, or remain unconnected.
Synthetic data generation pseudocode and hyperparameters 
can be found in Appx~\ref{appx:synth_data_gen}.
The resulting body was
voxelized within a 64 $\times$ 64 $\times$ 64 
cartesian grid. 
Each voxel in the grid was assigned an integer label $i \in [1, K+2]$, 
where $i=1$ represents empty space, 
$i=2$ represents a soft tissue, 
and $i > 2$ was used to represent the (at most $K$) independent rigid bones. 
A one-hot encoding was used to represent the workspace: a tensor with shape 
$(64, 64, 64, k+2)$.

A variational autoencoder (VAE; \cite{kingma2013auto}) 
with four blocks of Voxception-ResNet modules \citep{brock2016generative} 
was then trained 
to map voxel space into 
a highly compressed latent distribution
consisting of 512 latent dimensions.
Prior work that used voxel-based 
autoencoders considered fixed dataset, which limited their depth
and required data augmentation to avoid overfitting \citep{brock2016generative}. 
Because we can generate new synthetic data points on demand, our training data is unlimited, and the depth of our autoencoder---and thus its potential for compress, generalize and capture complex hierarchical features---is not constrained by lack of data.

Each design decoded from the latent space was automatically jointified (as detailed in Fig.~\ref{fig:joints}), and any exposed bone was covered with a thin layer of soft tissue. 
This tissue patching process was highly targeted; in randomly generated designs, patching was applied to less than 3\% of the workspace and covered less than 13\% of the robots' bodies. 
See Appx.~\ref{appx:more_hypers} for VAE hyperparameters.

\subsection{Co-Optimizing Endoskeletal Morphology and Control}
\label{sec:codesign}

\textbf{State space.}
Endoskeletal robots receive 
high dimensional 
mechanosensory data $\mathbf{s}^v = \{(x^v, y^v, z^v, \epsilon^v, \mathbf{v}^v), \dots \; | \; v \in \text{soft}\}$ with 3D position $x, y, z$, strain $\epsilon$ and velocity $\mathbf{v}$
from a variable number of soft voxels
that are distributed at variable locations in 3D space (Fig.~\ref{fig:RL}C), 
depending on the size, shape and deformation of its soft tissues.
In addition to sensing the movement and distortion of the soft tissues, proprioception of each rigid bone $r$  and joint $j$ between bones is also taken as input. 
The state of bones can be represented as $\mathbf{s}^r = \{(x^r, y^r, z^r, m^r, \mathbf{q}^r, \mathbf{v}^r, \boldsymbol{\omega}^r), \dots \; | \; r \in \text{bone}\}$ where $x, y, z$ is the position of the center of mass of the bone, $m$ is the mass, $\mathbf{q}$ is quaternion representing the orientation, 
and $\mathbf{v}$ and $\boldsymbol{\omega}$ are the linear velocity and angular velocity. 
The state of joints can be represented as $\mathbf{s}^j = \{(\mathbf{h}^j, \mathbf{d}_1^j, 
\mathbf{d}_2^j, \theta^j, \theta_{\text{min}}^j, \theta_{\text{max}}^j), \dots \; | \; j \in \text{joints}\}$ 
where $\mathbf{h}$ is the hinge axis direction of joint, 
$\mathbf{d}_1$ is the vector from the joint position to the center of mass of the first connected bone, $\mathbf{d}_2$ is the vector from the joint position to the center of mass of the second connected bone, 
$\theta^j$ is the current joint angle, and
$\theta_{\text{min}}^j$ and $\theta_{\text{max}}^j$ are the minimum and maximum joint angle limits, respectively.
Soft voxel sensing is
locally pooled 
into nodes 
at the center of mass of each bone (Fig.~\ref{fig:RL}D)
thus aligning 
soft tissue and skeletal
inputs 
(Fig.~\ref{fig:RL}E)
and promoting 
``reflex partitioning''
\citep{Windhorst_Hamm_Stuart_1989} 
whereby motors are affected 
more by local sensors
than distant sensors. 

\textbf{Action space.}
The action space consists of joint angles, 
which control the movements and articulation of the robot's internal skeleton. 
We use a uniform discrete angle action space $\mathcal{A} = \{ \text{-1.4 rad, -0.7 rad, 0 rad, 0.7 rad, 1.4 rad} \}$ for all joints. 
We selected 1.4 rad as the limit for joint rotation range 
to prevent extremely large unrealistic deformation of the soft materials surrounding the joints \citep{Bonet_Wood_2008}. 
We found 
that a discrete action space 
greatly simplified and stabilized 
policy training 
compared to an otherwise equivalent
continuous action space 
(Fig.~\ref{fig:discrete_vs_continous}).

\textbf{Reward.}
Reward was based on net displacement of the robot across 
100 pairs of observations and actions sampled at 10Hz during a simulation episode of 10 seconds. 
Rewards were generated per episode in the following way:
\begin{eqnarray}\label{eq:reward_all}
\mathbf{u}_{t} &=& (x_{t-1} - x_{0}, y_{t-1} - y_{0}) \\
\mathbf{\hat{u}}_{t} &=& \frac{\mathbf{u}_{t}}{||\mathbf{u}_{t}||} \\
\mathbf{v}_{t} &=& (x_{t} - x_{t-1}, y_{t} - y_{t-1}) \\
r_t &=&
\begin{cases}
    -P_{\text{large}}, & \text{if } ||\mathbf{v}_{t}|| < \delta \\
    \max(\mathbf{\hat{u}}_{t} \cdot \mathbf{v}_{t}, -P_{\text{small}}), & \text{otherwise}
\end{cases}\label{eq:reward_r}
\end{eqnarray}
where $x_t$ and $y_t$ are the x and y position of the center of mass of the robot at time step $t$,
$r_t$ the reward received by the robot, $\delta$ is a small threshold value for detecting stationary,
and $\mathbf{u}_t$ is the history movement vector up to the previous frame projected to the xy plane. 
If the norm of $\mathbf{u}_t$ is 0, 
we replace $\mathbf{\hat{u}}_{t}$ with vector $(1, 0)$ pointing in the positive X direction.
We apply a large penalty if robot is static and clip a negative reward to a smaller penalty if the robot is moving but in an undesired direction. 
Eq.~\ref{eq:reward_r} encourages the robot to maintain consistent forward motion while penalizing stagnation or drastic directional changes, 
which was found to be more robust for identifying designs with high locomotive ability 
compared to using a naive reward based on moving distance in a fixed direction, such as $r_t = x_t - x_{t-1}$. 
This is because the naive reward may discard designs with high locomotive ability in the wrong direction.
In some experiments the reward function was augmented to 
encourage robots to lift themselves above the surface plane using legs.

\textbf{Policy.} We employed Proximal Policy Optimization (PPO; \cite{schulman2017proximal}) 
to train a single universal controller for an evolving population of 
64
endoskeletal robots. 
%
A clone of each design in the population was created, yielding a batch of 128 designs.
This last step is intended to broaden the evaluation of each design and so reduce the likelihood of erroneously discarding a good design.
At each time step \(t\), the policy's observation \(\mathbf{s}_t = (\mathbf{s}^v_t, \mathbf{s}^j_t, \mathbf{s}^r_t)\) consists of soft voxel observations \(\mathbf{s}^v_t\), joint observations \(\mathbf{s}^j_t\), and rigid bone observations \(\mathbf{s}^r_t\).
Given these observations as input,
the policy 
returns
actions and value estimates according to the following equations:
\begin{align}
    \mathbf{s}^l_t &= \text{SpatialPool}(\mathbf{s}^v_t) \label{eq:spatial_pooling}\\
    \mathbf{n}_t &= [\mathbf{s}^r_t, \mathbf{s}^l_t] \label{eq:node_features} \\
    (\tilde{\mathbf{n}}_t, \tilde{\mathbf{e}}_t) &= \text{GraphTransformer}(\mathbf{n}_t, \mathbf{s}^j_t) \label{eq:graph_transformer} \\
    \mathbf{n}_t^* &= \text{MaxPool}(\tilde{\mathbf{n}}_t) \label{eq:max_pooling} \\
    V(\mathbf{s}_t) &= \text{ResNet}_V(\mathbf{n}_t^*) \label{eq:critic_value_function} \\
    \mathbf{z}_{t,j} &= \text{ResNet}_{\pi}(\tilde{\mathbf{e}}_{t,j}) \label{eq:actor_logits} \\
    \pi(a_{t,j} | \mathbf{s}_t) &= \text{Softmax}(\mathbf{z}_{t,j}) \label{eq:action_distribution} \\
    \pi(a_t | \mathbf{s}_t) &= \prod_{j} \pi(a_{t,j} | \mathbf{s}_t) \label{eq:overall_policy}
\end{align}
Soft voxel observations \(\mathbf{s}^v_t\) are locally pooled 
\citep{peng2020convolutional}
into a $64^3$ grid using the robot center of mass as the grid origin. 
The convoluted feature grid is then queried at the center of masses of each rigid bone, forming locally pooled node features  \(\mathbf{s}^l_t\) (Fig.~\ref{fig:RL}D).
This operation aligns soft tissue sensory inputs with skeletal sensing. 
The pooled features are then
concatenated with the rigid bone observations \(\mathbf{s}^r_t\) to form the node features \(\mathbf{n}_t\), as shown in Fig.~\ref{fig:RL}E.

A graph transformer \citep{shi2020masked} 
processes the robot's topology graph, taking the node features \(\mathbf{n}_t\) and edge features \(\mathbf{s}^j_t\) (joint observations) as input, 
and outputting processed node features \(\tilde{\mathbf{n}}_t\) 
and processed edge features \(\tilde{\mathbf{e}}_t\). 
The processed edge features are 
produced by concatenating
processed node features 
across the two nodes 
connected by each edge.
By max pooling (Fig.~\ref{fig:RL}L) 
over the processed node features \(\tilde{\mathbf{n}}_t\) 
we obtain a global feature vector \(\mathbf{n}_t^*\) (Fig.~\ref{fig:RL}M), 
which summarizes the overall state of the robot.
The Critic (Fig.~\ref{fig:RL}N) uses \(\text{ResNet}_V\) 
to take the pooled node feature \(\mathbf{n}_t^*\) and output the value function \(V(\mathbf{s}_t)\). 
The Actor (Fig.~\ref{fig:RL}I) processes edge features \(\tilde{\mathbf{e}}_{t,j}\), corresponding to each joint \(j\), and passes them through \(\text{ResNet}_{\pi}\) to compute the action logits \(\mathbf{z}_{t,j}\) which define
the action distribution for joint \(j\). 
By computing the softmax over the logits \(\mathbf{z}_{t,j}\), 
probabilities are obtained over the discrete action space $\mathcal{A}$.
The overall policy \(\pi(a_t | \mathbf{s}_t)\) was thus 
defined as the product 
of the action distributions 
over all joints, 
assuming independence between joints.

By utilizing a graph transformer to process the robot's topology, 
the policy effectively learns to condition on 
both the sensory input and the morphology of a robot. 
This architecture allows the actor to generate joint-specific actions while the critic evaluates the discounted score $V(\mathbf{s}_t) = \mathbb{E}_{\pi} \left[ \sum_{t=0}^{T} \gamma^t r_t \right]$.

\textbf{Fitness.}
The aggregate behavior of a robot 
across an entire simulation episode
was evaluated against a fitness function.
After 20 epochs of learning, 
each design in the current population was assigned a fitness score equal to the peak fitness it achieved across 40 episodes of simulation.

\textbf{Evolutionary design.}
The population of designs 
fed to the universal controller 
was optimized by covariance matrix evolutionary strategies (CMA-ES; 
\cite{hansen2001completely}). 
Briefly,
a multivariate normal distribution of designs is sampled from the latent space
and 
the mean vector is ``pulled'' toward the latent coordinates of sampled designs with the highest fitness 
and the covariance matrix is adapted to stretch along the most promising dimensions and shrink within others.
Because CMA-ES 
uses ranked fitness scores 
instead of raw fitness scores
to update the mean design vector,
undesirable designs
can be assigned 
a large negative rank 
without interfering with the stability of the optimizer.
Body plans that have 
less than two joints 
or less than 20\% bone,
were  
deemed ``invalid designs'' 
and assigned a 
large negative reward 
as described below in Sect.~\ref{sec:codesign}.
Less than 15\% of the randomly generated robots were invalid under this definition of validity.

\begin{figure}[t]
    \centering
    \includegraphics[width=\textwidth]{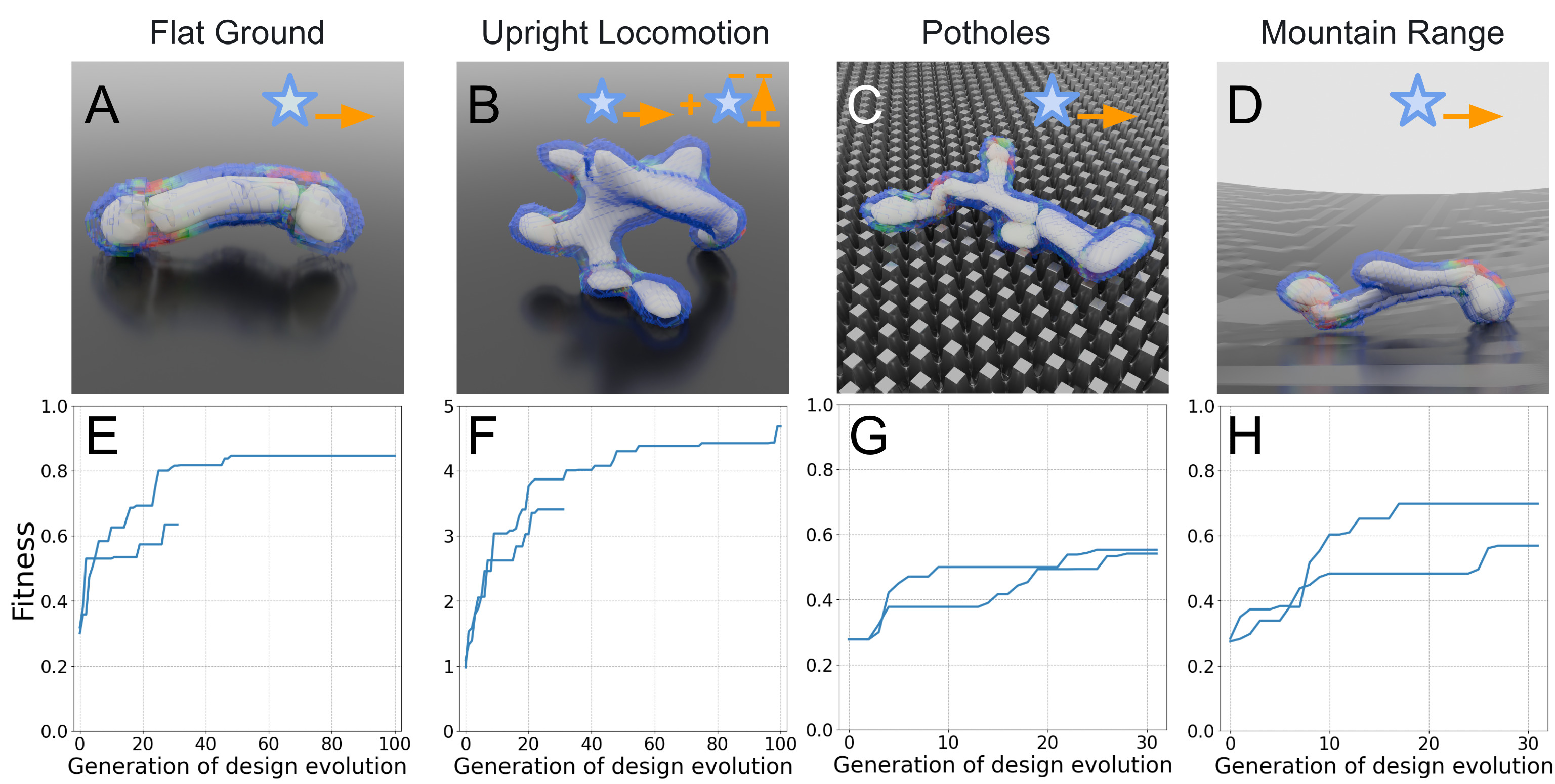}
    \vspace{-12pt}
    
    \caption{\textbf{Task environments.}
    We considered four task environments: 
    Flat Ground (\textbf{A}),
    Upright Locomotion (on flat ground; \textbf{B}),
    Potholes (\textbf{C})
    and
    Mountain Range (\textbf{D}).
    In each one,
    two independent evolutionary trials were conducted,
    and
    the peak fitness achieved 
    by each design was
    averaged across 
    the population
    before plotting the cumulative max
    (\textbf{E-H}).
    The best design is shown for each environment (A-D).
    Rewarding net displacement over flat ground (A,E) results in the evolution of snakes (A).
    We ran optimization for extra time (100 generations) to see if legs would emerge, but evolution
    failed to escape the local optima of snakes.
    Adding a second component to the reward function 
    that tracks the proportion of body voxels
    that fall below a prespecified height threshold during behavior
    (B,F) resulted in legged locomotion (B),
    and continued to innovate when provided extra time.
    Increasing the difficultly of the terrain also promoted morphological diversity (C,D).
    \vspace{-8pt}
}
    \label{fig:env}
\end{figure}

\begin{figure}[t]
    \centering
    \includegraphics[width=\textwidth]{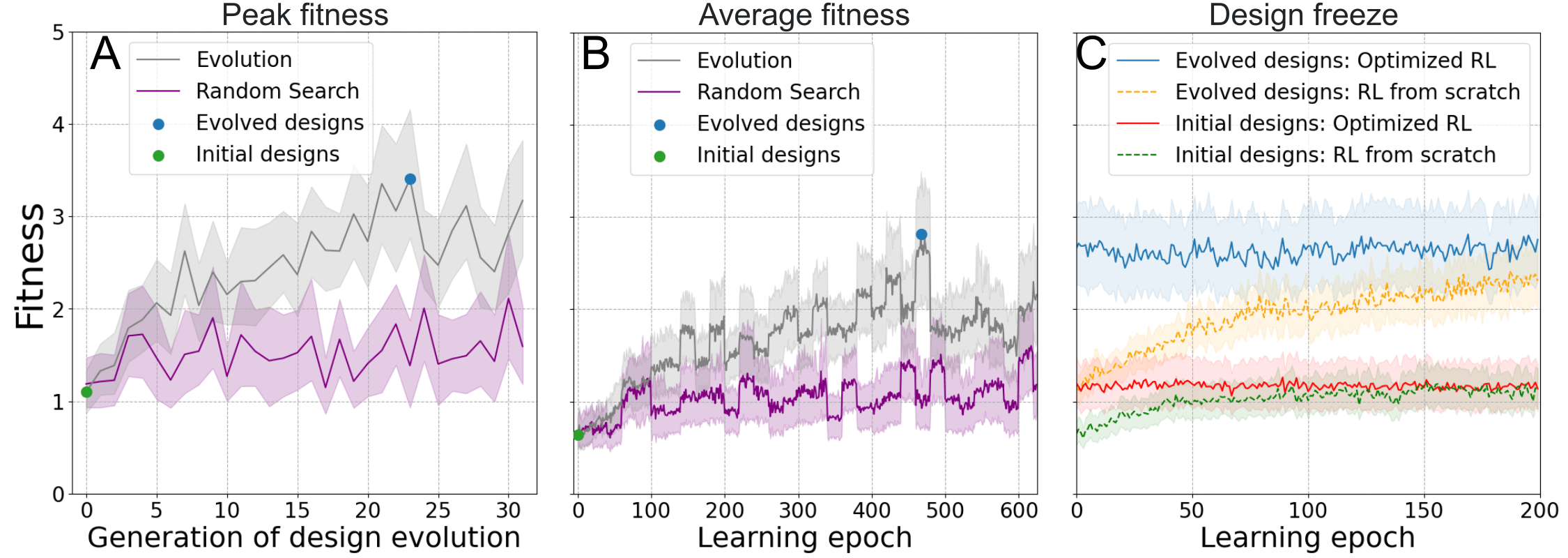}
    \vspace{-16pt}
    
    \caption{\textbf{Evolution of learning.}
    A population of 64 designs was 
    iteratively drawn from an evolving design distribution and
    used by reinforcement learning to obtain a universal controller.
    The peak fitness achieved 
    by each design 
    across 20 epochs of RL
    is plotted (with 95\% bootstrapped confidence intervals)
    at each generation of design evolution
    in the Upright Locomotion task environment (\textbf{A}).
    The dark gray line in A is mean peak fitness across the 64 designs in the population; the shaded region is the CI of the mean.
    Similarly,
    the average fitness achieved by each design 
    across its simulation episodes
    is shown
    at each learning epoch
    (in \textbf{B}).
    An otherwise equivalent experiment 
    with random morphological search instead of evolution strategies was also performed (purple curves in A and B).
    After 30 generations of evolution (600 total epochs of RL), 
    the 64 designs from the 
    best evolved population
    (blue dot; gen 23) 
    were compared 
    to the
    64 designs from
    initial randomly-generated population (green dot; gen 0).
    To do so, the designs within each population were frozen and their behavior was re-optimized (\textbf{C})
    starting with the optimized universal controller from gen 23 (Optimized RL). 
    A second comparison was then made in which a new universal controller 
    re-learned
    for each population separately, 
    from scratch
    (RL from scratch). 
    In both comparisons, the learning conditions were equivalent and the only differences between the populations were morphological.
    And in both comparisons the evolved designs easily outperformed the initial designs. 
    This 
    indicates that design evolution 
    was necessary:
    it pushed search into regions of the
    latent space containing better morphologies
    with higher fitness potential
    that better 
    support universal controller training.
    \vspace{-8pt}
}
    \label{fig:before-after-evo}
\end{figure}

\section{Results}
\label{sec:results}

In this section, 
we present our learned 
endoskeletal latent embedding
and its use in 
body-brain co-design 
across four different task environments.
We also perform a series of control experiments that isolate the effect of morphological evolution in order to determine its role in achieving high performing endoskeletal robots.

\subsection{Endoskeletal Latent Space}

The learned latent space is 
smooth and expressive (Fig.~\ref{fig:latent})
and was found to contain tunable genes that encode 
specific morphological traits, such as body height, length, width, volume and bone count (Figs.~\ref{fig:latent_correlation}-\ref{fig:gene_editing}).
The trained VAE is able 
to encode and decode 
previously unseen 
training data points 
with high accuracy (Fig.~\ref{fig:synth_vs_decode})
while generalizing beyond the training data:
Most designs sampled from the latent space
look very different from the multi-star graphs (see e.g.~Fig.~\ref{fig:synth_vs_decode}A) used to train the VAE.
For example, 
skeletal voids 
(i.e.~holes that pass through individual bones) 
were not present in the generated synthetic training data set
but 
they are present in the bones of
decoded designs from randomly-sampled latent codes (Fig.~\ref{fig:latent}A$'$).

\subsection{Task Environments}

\textbf{Flat Ground.}
On perfectly flat ground, selection for forward locomotion 
resulted in the evolution of
snake-like body plans with two joints (Fig.~\ref{fig:env}A) that are able to move at high speeds.

\textbf{Upright Locomotion.}
Given the impressive speed of snakes on flat ground, a second term was added to the reward function to  
promote body plans that exhibit upright legged locomotion. 
To do so,
the proportion of body voxels that fell within 5 cm of the surface plane during behavior
was subtracted from the original reward at each step of learning,
and the 
original fitness was divided by the 
mean proportion of voxels that fell below this threshold across the full simulation episode.
This resulted in
less serpentine body plans with more joints (Fig.~\ref{fig:standing_grid_evolved}).
The best evolved design is much more complex, and uses its legs to lift its body above the ground during locomotion (Fig.~\ref{fig:env}B).

\textbf{Potholes.} 
Next, we explored a 
more challenging terrain.
In the Potholes environment,
there are regularly-spaced depressions (``potholes'')
along the surface plane.
The depth, width, and spacing of the potholes can be easily adjusted 
in the provided code
(indeed any 
2D height matrix 
can be supplied 
to change the terrain).
With relatively small,
tightly-spaced 
potholes,
the best evolved designs 
(Fig.~\ref{fig:env}C) 
are
longer and more complex
than the snakes that
dominated 
flat ground
under the same 
fitness function
(Fig.~\ref{fig:env}A).
A common adaptation that evolved for navigating potholes was broad feet, which the robots used to
push off 
and pull
against
the
vertical walls inside the potholes.

\textbf{Mountain Range.} 
In our fourth and final environment,
robots must climb up a slope in order to move away from the origin.
In this mountain environment, the best evolved designs (Fig.~\ref{fig:env}D) were also slightly more complex than the flatland snakes.


\subsection{The Necessity of Morphological Evolution}

Across all four task environments,
evolved design populations significantly outperform 
the randomly-generated initial populations of designs 
from the first generation of evolution. 
This could be due to improvements in the universal controller or better body plans, or both.
In order to isolate the effects of morphological evolution and policy training, we performed the following three
control experiments using the Upright Locomotion task environment. 

In the first experiment, we extracted 
the design population that achieved the highest mean reward during evolution (Evolved designs in Fig.~\ref{fig:before-after-evo}B),
as well as
the RL model checkpoint at this point (Optimized RL in Fig.~\ref{fig:before-after-evo}C). 
We also extracted the initial designs that were randomly generated at the very beginning of evolution (gen 0).
The initial designs (Fig.~\ref{fig:standing_grid_init}) were frozen and the controller was re-optimized using this population of frozen designs for 100 epochs starting from optimized RL checkpoint.
In a parallel independent experiment, the evolved designs (Fig.~\ref{fig:standing_grid_evolved}) were likewise frozen and the controller was re-optimized using the frozen evolved population of designs for 100 epochs starting from optimized RL checkpoint.
The same procedure was repeated, this time retraining the control policy from scratch, for both frozen design populations separately.
In both cases---starting policy training from scratch or from the RL checkpoint---the frozen evolved designs significantly outperformed the frozen initial designs (Fig.~\ref{fig:before-after-evo}C).
In both frozen design populations, 
the universal controller from the checkpoint
immediately exhibits the best performance achieved after retraining from scratch.
This shows that RL does not suffer catastrophic forgetting: 
the policy does not forget how to control the initial designs as the population evolves to contain different, better designs.

These results suggest that morphological evolution yields better designs but they do not prove that evolution is necessary.
It could be the case that good designs will simply arise by random mutations alone, without cumulative selection across discrete generations.
As a final control experiment we tested this hypothesis by replacing evolutionary strategies with random search
(purple curve in Fig.~\ref{fig:before-after-evo}A,B).
Random morphological search performed significantly worse than evolutionary strategies.
This suggests that morphological evolution (mutation \textit{and} selection) is indeed necessary.

\section{Discussion}
\label{sec:discussion}

In this paper we
introduced the 
computational design of
freeform endoskeletal robots.
The open-ended nature of the design problem allowed us to
procedurally generate a never-ending synthetic data flow 
of endoskeletal training examples 
for representation learning, 
which in turn allowed 
for effective learning of 
a much deeper voxel-based autoencoder 
than 
those learned in prior work with fixed datasets \citep{brock2016generative}.
This resulted in a very smooth 
and highly expressive latent embedding 
of design space (Fig.~\ref{fig:latent}),
which we employed  
as a genotype space 
(and the decoder as the genotype-phenotype mapping) 
for morphological evolution.

We found that a universal controller could be simultaneously obtained 
by reinforcement learning
during morphological evolution
despite the 
wide range 
of endoskeletal geometries and topologies 
that emerged within the evolving population.
We observed across several independent experimental trials that morphological evolution tended to push search into promising 
regions of latent space consisting of high performing body plans that facilitated policy training.
For flat ground, this resulted in the evolution of simple two-jointed 
snakes (Fig.~\ref{fig:env}A), 
which in hindsight makes intuitive sense:
such bodies are easier to control.
But, in the other three task environments we tested,
many other, more complex solutions evolved, 
including legged locomotion (Fig.~\ref{fig:env}B).
This suggests that other researchers may leverage the open-endedness of
this computational design platform
to 
identify
which
physical environments and 
fitness pressures 
lead to the emergence of particular abilities 
and the embodied structures required to support those abilities
 
Others may use our simulator as 
an easy-to-integrate 
software library
for
benchmarking 
their own voxel-based representations and learning algorithms
in the four task environments released here, 
or by creating their own terrain maps and reward functions based on these examples.
To this end, 
we also provide
an example of
object manipulation
(``dribbling a football''; Fig.~\ref{fig:soccer}),
which we hope inspires others to 
create their own objects 
and
build more, and more intricate, virtual worlds.

However,
there were key limitations 
of the present work that may be overcome by 
future work.
First, 
because skeletons were assumed to be inside of each robot, rigid body collisions were not modeled,
and thus
external rigid horns, claws, shells, etc.~were not possible.
Also,
fluids were not modeled,
and thus
the simulator was
restricted to
terrestrial environments and land based behaviors.
But the most important limitation
of this paper 
is that
simulated designs were not realized as physical robots.
Recent advances in volumetric 3D printing
\citep{darkes2024volumetric}
are 
beginning to enable the 
fabrication
of
endoskeletal bodies 
that 
flex their muscles 
but are not yet capable of locomotion.
This suggests that
artificial endoskeletal systems
will soon move through our world,
opening the way to
evolved endoskeletal robots
that begin to
approach the sophisticated---%
and perhaps eventually cognitive behavior of 
evolved endoskeletal organisms.

\subsubsection*{Acknowledgments}

This research was supported by
NSF award FRR-2331581,
Schmidt Sciences AI2050 grant G-22-64506,
Templeton World Charity Foundation award no.~20650,
and the Berggruen Institute. 

\bibliography{main}

\begin{thebibliography}{46}
\providecommand{\natexlab}[1]{#1}
\providecommand{\url}[1]{\texttt{#1}}
\expandafter\ifx\csname urlstyle\endcsname\relax
  \providecommand{\doi}[1]{doi: #1}\else
  \providecommand{\doi}{doi: \begingroup \urlstyle{rm}\Url}\fi

\bibitem[Auerbach \& Bongard(2014)Auerbach and Bongard]{auerbach2014environmental}
Joshua~E Auerbach and Josh~C Bongard.
\newblock Environmental influence on the evolution of morphological complexity in machines.
\newblock \emph{PLoS Computational Biology}, 10\penalty0 (1):\penalty0 e1003399, 2014.

\bibitem[Baraff(2001)]{baraff2001physically}
David Baraff.
\newblock Physically based modeling: Rigid body simulation.
\newblock \emph{SIGGRAPH Course Notes}, 2\penalty0 (1):\penalty0 2--1, 2001.

\bibitem[Baumgarte(1972)]{baumgarte1972stabilization}
Joachim Baumgarte.
\newblock Stabilization of constraints and integrals of motion in dynamical systems.
\newblock \emph{Computer Methods in Applied Mechanics and Engineering}, 1\penalty0 (1):\penalty0 1--16, 1972.

\bibitem[Bonet \& Wood(2008)Bonet and Wood]{Bonet_Wood_2008}
Javier Bonet and Richard~D. Wood.
\newblock \emph{Nonlinear Continuum Mechanics for Finite Element Analysis}.
\newblock Cambridge University Press, 2nd edition, 2008.

\bibitem[Brock et~al.(2016)Brock, Lim, Ritchie, and Weston]{brock2016generative}
Andrew Brock, Theodore Lim, James~M Ritchie, and Nick Weston.
\newblock Generative and discriminative voxel modeling with convolutional neural networks.
\newblock \emph{arXiv preprint arXiv:1608.04236}, 2016.

\bibitem[Cheney et~al.(2013)Cheney, MacCurdy, Clune, and Lipson]{cheney2013unshackling}
Nick Cheney, Robert MacCurdy, Jeff Clune, and Hod Lipson.
\newblock Unshackling evolution: Evolving soft robots with multiple materials and a powerful generative encoding.
\newblock In \emph{Proceedings of the Conference on Genetic and Evolutionary Computation (GECCO)}, pp.\  167--174. ACM, 2013.

\bibitem[Cheney et~al.(2018)Cheney, Bongard, SunSpiral, and Lipson]{cheney2018scalable}
Nick Cheney, Josh Bongard, Vytas SunSpiral, and Hod Lipson.
\newblock Scalable co-optimization of morphology and control in embodied machines.
\newblock \emph{Journal of The Royal Society Interface}, 15\penalty0 (143):\penalty0 20170937, 2018.

\bibitem[Cochevelou et~al.(2023)Cochevelou, Bonner, and Schmidt]{cochevelou2023differentiable}
Fran\c{c}ois Cochevelou, David Bonner, and Martin-Pierre Schmidt.
\newblock Differentiable soft-robot generation.
\newblock In \emph{Proceedings of the Genetic and Evolutionary Computation Conference (GECCO)}, pp.\  129–137. ACM, 2023.

\bibitem[Darkes-Burkey \& Shepherd(2024)Darkes-Burkey and Shepherd]{darkes2024volumetric}
Cameron Darkes-Burkey and Robert~F Shepherd.
\newblock Volumetric {3D} printing of endoskeletal soft robots.
\newblock \emph{Advanced Materials}, pp.\  2402217, 2024.

\bibitem[Erleben(2013)]{erleben2013numerical}
Kenny Erleben.
\newblock Numerical methods for linear complementarity problems in physics-based animation.
\newblock In \emph{ACM SIGGRAPH 2013 Courses}, pp.\  1--42, 2013.

\bibitem[Guimar{\~a}es et~al.(2020)Guimar{\~a}es, Gasperini, Marques, and Reis]{guimaraes2020stiffness}
Carlos~F Guimar{\~a}es, Luca Gasperini, Alexandra~P Marques, and Rui~L Reis.
\newblock The stiffness of living tissues and its implications for tissue engineering.
\newblock \emph{Nature Reviews Materials}, 5\penalty0 (5):\penalty0 351--370, 2020.

\bibitem[Gupta et~al.(2021)Gupta, Savarese, Ganguli, and Fei-Fei]{gupta2021embodied}
Agrim Gupta, Silvio Savarese, Surya Ganguli, and Li~Fei-Fei.
\newblock Embodied intelligence via learning and evolution.
\newblock \emph{Nature Communications}, 12\penalty0 (1):\penalty0 1--12, 2021.

\bibitem[Gupta et~al.(2022)Gupta, Fan, Ganguli, and Fei-Fei]{gupta2022metamorph}
Agrim Gupta, Linxi Fan, Surya Ganguli, and Li~Fei-Fei.
\newblock Metamorph: Learning universal controllers with transformers.
\newblock In \emph{International Conference on Learning Representations (ICLR)}, 2022.

\bibitem[Hansen \& Ostermeier(2001)Hansen and Ostermeier]{hansen2001completely}
Nikolaus Hansen and Andreas Ostermeier.
\newblock Completely derandomized self-adaptation in evolution strategies.
\newblock \emph{Evolutionary Computation}, 9\penalty0 (2):\penalty0 159--195, 2001.

\bibitem[Hejna et~al.(2021)Hejna, Abbeel, and Pinto]{iii2021taskagnostic}
Donald~J Hejna, Pieter Abbeel, and Lerrel Pinto.
\newblock Task-agnostic morphology evolution.
\newblock In \emph{Proceedings of the International Conference on Learning Representations (ICLR)}, 2021.

\bibitem[Hiller \& Lipson(2010)Hiller and Lipson]{hiller2010evolving}
Jonathan Hiller and Hod Lipson.
\newblock Evolving amorphous robots.
\newblock In \emph{Proceedings of the Conference on Artificial Life (ALife)}, pp.\  717--724, 2010.

\bibitem[Hiller \& Lipson(2012)Hiller and Lipson]{hiller2012automatic}
Jonathan Hiller and Hod Lipson.
\newblock Automatic design and manufacture of soft robots.
\newblock \emph{IEEE Transactions on Robotics}, 28\penalty0 (2):\penalty0 457--466, 2012.

\bibitem[Hiller \& Lipson(2014)Hiller and Lipson]{hiller2014dynamic}
Jonathan Hiller and Hod Lipson.
\newblock Dynamic simulation of soft multimaterial 3{D}-printed objects.
\newblock \emph{Soft Robotics}, 1\penalty0 (1):\penalty0 88--101, 2014.

\bibitem[Hu et~al.(2023)Hu, Whitman, and Choset]{hu2023glso}
Jiaheng Hu, Julian Whitman, and Howie Choset.
\newblock {GLSO: G}rammar-guided latent space optimization for sample-efficient robot design automation.
\newblock In \emph{Proccedings of the Conference on Robot Learning (CoRL)}, pp.\  1321--1331. PMLR, 2023.

\bibitem[Huang et~al.(2024)Huang, Chen, Xu, and Sitzmann]{huang2024dittogym}
Suning Huang, Boyuan Chen, Huazhe Xu, and Vincent Sitzmann.
\newblock Dittogym: Learning to control soft shape-shifting robots.
\newblock In \emph{Proceedings of the International Conference on Learning Representations (ICLR)}, 2024.

\bibitem[Jakobi et~al.(1995)Jakobi, Husbands, and Harvey]{jakobi1995noise}
Nick Jakobi, Phil Husbands, and Inman Harvey.
\newblock Noise and the reality gap: The use of simulation in evolutionary robotics.
\newblock In \emph{Proceedings of the European Conference on Artificial Life (ECAL)}, pp.\  704--720, 1995.

\bibitem[Jansson \& Vergeest(2003)Jansson and Vergeest]{jansson2003combining}
Johan Jansson and Joris~SM Vergeest.
\newblock Combining deformable-and rigid-body mechanics simulation.
\newblock \emph{The Visual Computer}, 19:\penalty0 280--290, 2003.

\bibitem[Kim \& Pollard(2011)Kim and Pollard]{kim2011fast}
Junggon Kim and Nancy~S Pollard.
\newblock Fast simulation of skeleton-driven deformable body characters.
\newblock \emph{ACM Transactions on Graphics (TOG)}, 30\penalty0 (5):\penalty0 1--19, 2011.

\bibitem[Kingma \& Welling(2013)Kingma and Welling]{kingma2013auto}
Diederik~P Kingma and Max Welling.
\newblock Auto-encoding variational bayes.
\newblock \emph{arXiv preprint arXiv:1312.6114}, 2013.

\bibitem[Kriegman et~al.(2020{\natexlab{a}})Kriegman, Blackiston, Levin, and Bongard]{kriegman2020xenobots}
Sam Kriegman, Douglas Blackiston, Michael Levin, and Josh Bongard.
\newblock A scalable pipeline for designing reconfigurable organisms.
\newblock \emph{Proceedings of the National Academy of Sciences}, 117\penalty0 (4):\penalty0 1853--1859, 2020{\natexlab{a}}.

\bibitem[Kriegman et~al.(2020{\natexlab{b}})Kriegman, Nasab, Shah, Steele, Branin, Levin, Bongard, and Kramer-Bottiglio]{kriegman2020scalable}
Sam Kriegman, Amir~Mohammadi Nasab, Dylan Shah, Hannah Steele, Gabrielle Branin, Michael Levin, Josh Bongard, and Rebecca Kramer-Bottiglio.
\newblock Scalable sim-to-real transfer of soft robot designs.
\newblock In \emph{Proceedings of the International Conference on Soft Robotics (RoboSoft)}, pp.\  359--366, 2020{\natexlab{b}}.

\bibitem[Kriegman et~al.(2021)Kriegman, Blackiston, Levin, and Bongard]{kriegman2021kinematic}
Sam Kriegman, Douglas Blackiston, Michael Levin, and Josh Bongard.
\newblock Kinematic self-replication in reconfigurable organisms.
\newblock \emph{Proceedings of the National Academy of Sciences}, 118\penalty0 (49):\penalty0 e2112672118, 2021.

\bibitem[Li et~al.(2020)Li, Liu, and Kavan]{li2020soft}
Jing Li, Tiantian Liu, and Ladislav Kavan.
\newblock Soft articulated characters in projective dynamics.
\newblock \emph{IEEE Transactions on Visualization and Computer Graphics}, 28\penalty0 (2):\penalty0 1385--1396, 2020.

\bibitem[Li et~al.(2024)Li, Matthews, and Kriegman]{li2024reinforcement}
Muhan Li, David Matthews, and Sam Kriegman.
\newblock Reinforcement learning for freeform robot design.
\newblock In \emph{Proceedings of the International Conference on Robotics and Automation (ICRA)}, 2024.

\bibitem[Lipson \& Pollack(2000)Lipson and Pollack]{lipson2000automatic}
Hod Lipson and Jordan~B Pollack.
\newblock Automatic design and manufacture of robotic lifeforms.
\newblock \emph{Nature}, 406\penalty0 (6799):\penalty0 974, 2000.

\bibitem[Lowrey et~al.(2019)Lowrey, Rajeswaran, Kakade, Todorov, and Mordatch]{lowrey2018plan}
Kendall Lowrey, Aravind Rajeswaran, Sham Kakade, Emanuel Todorov, and Igor Mordatch.
\newblock Plan online, learn offline: Efficient learning and exploration via model-based control.
\newblock In \emph{Proceedings of the International Conference on Learning Representations (ICLR)}, 2019.

\bibitem[Matthews et~al.(2023)Matthews, Spielberg, Rus, Kriegman, and Bongard]{matthews2023efficient}
David Matthews, Andrew Spielberg, Daniela Rus, Sam Kriegman, and Josh Bongard.
\newblock Efficient automatic design of robots.
\newblock \emph{Proceedings of the National Academy of Sciences}, 120\penalty0 (41):\penalty0 e2305180120, 2023.

\bibitem[Peng et~al.(2020)Peng, Niemeyer, Mescheder, Pollefeys, and Geiger]{peng2020convolutional}
Songyou Peng, Michael Niemeyer, Lars Mescheder, Marc Pollefeys, and Andreas Geiger.
\newblock Convolutional occupancy networks.
\newblock In \emph{Proceedings of the European Conference on Computer Vision (ECCV)}, pp.\  523--540. Springer, 2020.

\bibitem[Schulman et~al.(2017)Schulman, Wolski, Dhariwal, Radford, and Klimov]{schulman2017proximal}
John Schulman, Filip Wolski, Prafulla Dhariwal, Alec Radford, and Oleg Klimov.
\newblock Proximal policy optimization algorithms.
\newblock \emph{arXiv preprint arXiv:1707.06347}, 2017.

\bibitem[Shi et~al.(2020)Shi, Huang, Feng, Zhong, Wang, and Sun]{shi2020masked}
Yunsheng Shi, Zhengjie Huang, Shikun Feng, Hui Zhong, Wenjin Wang, and Yu~Sun.
\newblock Masked label prediction: Unified message passing model for semi-supervised classification.
\newblock \emph{arXiv preprint arXiv:2009.03509}, 2020.

\bibitem[Shinar et~al.(2008)Shinar, Schroeder, and Fedkiw]{shinar2008two}
Tamar Shinar, Craig Schroeder, and Ronald Fedkiw.
\newblock Two-way coupling of rigid and deformable bodies.
\newblock In \emph{Proceedings of the Symposium on Computer Animation (SCA)}, pp.\  95--103, 2008.

\bibitem[Sims(1994)]{sims1994competition}
Karl Sims.
\newblock Evolving 3{D} morphology and behavior by competition.
\newblock \emph{Artificial Life}, 1\penalty0 (4):\penalty0 353--372, 1994.

\bibitem[Stanley(2007)]{stanley2007compositional}
Kenneth~O Stanley.
\newblock Compositional pattern producing networks: A novel abstraction of development.
\newblock \emph{Genetic Programming and Evolvable Machines}, 8:\penalty0 131--162, 2007.

\bibitem[Strgar et~al.(2024)Strgar, Matthews, Hummer, and Kriegman]{strgar2024evolution}
Luke Strgar, David Matthews, Tyler Hummer, and Sam Kriegman.
\newblock Evolution and learning in differentiable robots.
\newblock In \emph{Robotics: Science and Systems (RSS)}, 2024.

\bibitem[Ventrella(1994)]{ventrella1994explorations}
Jeffrey Ventrella.
\newblock Explorations in the emergence of morphology and locomotion behavior in animated characters.
\newblock In \emph{Proceedings of the International Workshop on the Synthesis and Simulation of Living Systems (ALife)}, pp.\  436--441, 1994.

\bibitem[Wang et~al.(2023{\natexlab{a}})Wang, Ma, Spielberg, Xian, Zhang, Tenenbaum, Rus, and Gan]{wang2023softzoo}
Tsun-Hsuan Wang, Pingchuan Ma, Andrew~E. Spielberg, Zhou Xian, Hao Zhang, Joshua~B. Tenenbaum, Daniela Rus, and Chuang Gan.
\newblock Softzoo: A soft robot co-design benchmark for locomotion in diverse environments.
\newblock In \emph{Proceedings of the International Conference on Learning Representations (ICLR)}, 2023{\natexlab{a}}.

\bibitem[Wang et~al.(2023{\natexlab{b}})Wang, Zheng, Ma, Du, Kim, Spielberg, Tenenbaum, Gan, and Rus]{wang2023diffusebot}
Tsun-Hsuan Wang, Juntian Zheng, Pingchuan Ma, Yilun Du, Byungchul Kim, Andrew~E. Spielberg, Joshua~B. Tenenbaum, Chuang Gan, and Daniela Rus.
\newblock Diffusebot: Breeding soft robots with physics-augmented generative diffusion models.
\newblock In \emph{Advances in Neural Information Processing Systems (NeurIPS)}, 2023{\natexlab{b}}.

\bibitem[Windhorst et~al.(1989)Windhorst, Hamm, and Stuart]{Windhorst_Hamm_Stuart_1989}
Uwe Windhorst, Thomas~M. Hamm, and Douglas~G. Stuart.
\newblock On the function of muscle and reflex partitioning.
\newblock \emph{Behavioral and Brain Sciences}, 12\penalty0 (4):\penalty0 629–645, 1989.

\bibitem[Yuan et~al.(2022)Yuan, Song, Luo, Sun, and Kitani]{yuan2022transformact}
Ye~Yuan, Yuda Song, Zhengyi Luo, Wen Sun, and Kris Kitani.
\newblock Transform2{A}ct: Learning a transform-and-control policy for efficient agent design.
\newblock In \emph{Proceedings of the International Conference on Learning Representations (ICLR)}, 2022.

\bibitem[Yun et~al.(2019)Yun, Jeong, Kim, Kang, and Kim]{yun2019graph}
Seongjun Yun, Minbyul Jeong, Raehyun Kim, Jaewoo Kang, and Hyunwoo~J Kim.
\newblock Graph transformer networks.
\newblock In H.~Wallach, H.~Larochelle, A.~Beygelzimer, F.~d\textquotesingle Alch\'{e}-Buc, E.~Fox, and R.~Garnett (eds.), \emph{Advances in Neural Information Processing Systems}, volume~32. Curran Associates, Inc., 2019.

\bibitem[Zhao et~al.(2020)Zhao, Xu, Konakovi{\'c}-Lukovi{\'c}, Hughes, Spielberg, Rus, and Matusik]{zhao2020robogrammar}
Allan Zhao, Jie Xu, Mina Konakovi{\'c}-Lukovi{\'c}, Josephine Hughes, Andrew Spielberg, Daniela Rus, and Wojciech Matusik.
\newblock {RoboGrammar: G}raph grammar for terrain-optimized robot design.
\newblock \emph{ACM Transactions on Graphics (TOG)}, 39\penalty0 (6):\penalty0 1--16, 2020.

\end{thebibliography}
\bibliographystyle{iclr2025_conference}

\clearpage

\appendix

\section{Supplemental Figures}
\label{appx:figs}

\begin{figure}[h]
    \centering
    \includegraphics[width=\textwidth]{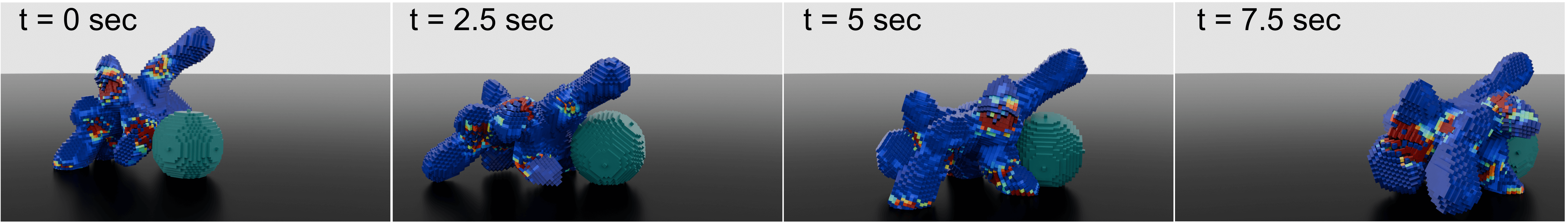}
    \vspace{-16pt}
    \caption{
    \textbf{Object manipulation.}
    One of the designs that evolved for Upright Locomotion performs a new task: it pushes an object forward.
    This design was among several others that we 
    sampled from the evolving population analyzed in Fig.~\ref{fig:before-after-evo},
    placed behind a soft sphere (teal),
    and actuated
    using the
    universal controller from the Optimized RL checkpoint in Fig.~\ref{fig:before-after-evo}C. 
    Videos of the resulting behaviors and the source code necessary to reproduce our results can be found on our project page.
    \href{https://endoskeletal.github.io}{\color{blue}https://endoskeletal.github.io}.
    }
    \label{fig:soccer}
\end{figure}

\vspace{2em}

\begin{figure}[h]
    \centering
    \includegraphics[width=\textwidth]{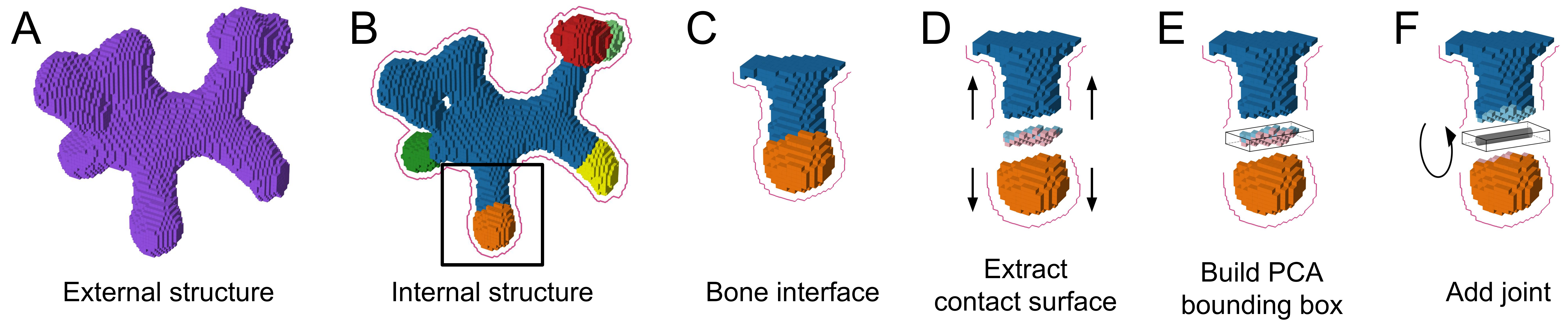}
    \vspace{-16pt}
    \caption{\textbf{Jointification.}
    The learned latent space
    encodes a generative model of an endoskeletal body plan:
    a external soft tissue geometry (\textbf{A})
    and an internally segmented skeleton (\textbf{B}).
    The latent space also implicitly encodes the location and orientation of each joint in the skeleton: 
    at the interface of each pair of bones in voxel space (\textbf{C}).
    In order to simulate a joint,
    voxels along the contact surface 
    of two opposing bones are extracted (\textbf{D})
    and Principal Component Analysis is used to construct an oriented bounded box around them (\textbf{E}).
    A hinge joint is then positioned at the center of mass of the contact surface with a direction along with the longest axis of the bounding box (\textbf{F}).
}
    \label{fig:joints}
\end{figure}

\vspace{2em}

\begin{figure}[h]
    \centering
    \includegraphics[width=\textwidth]{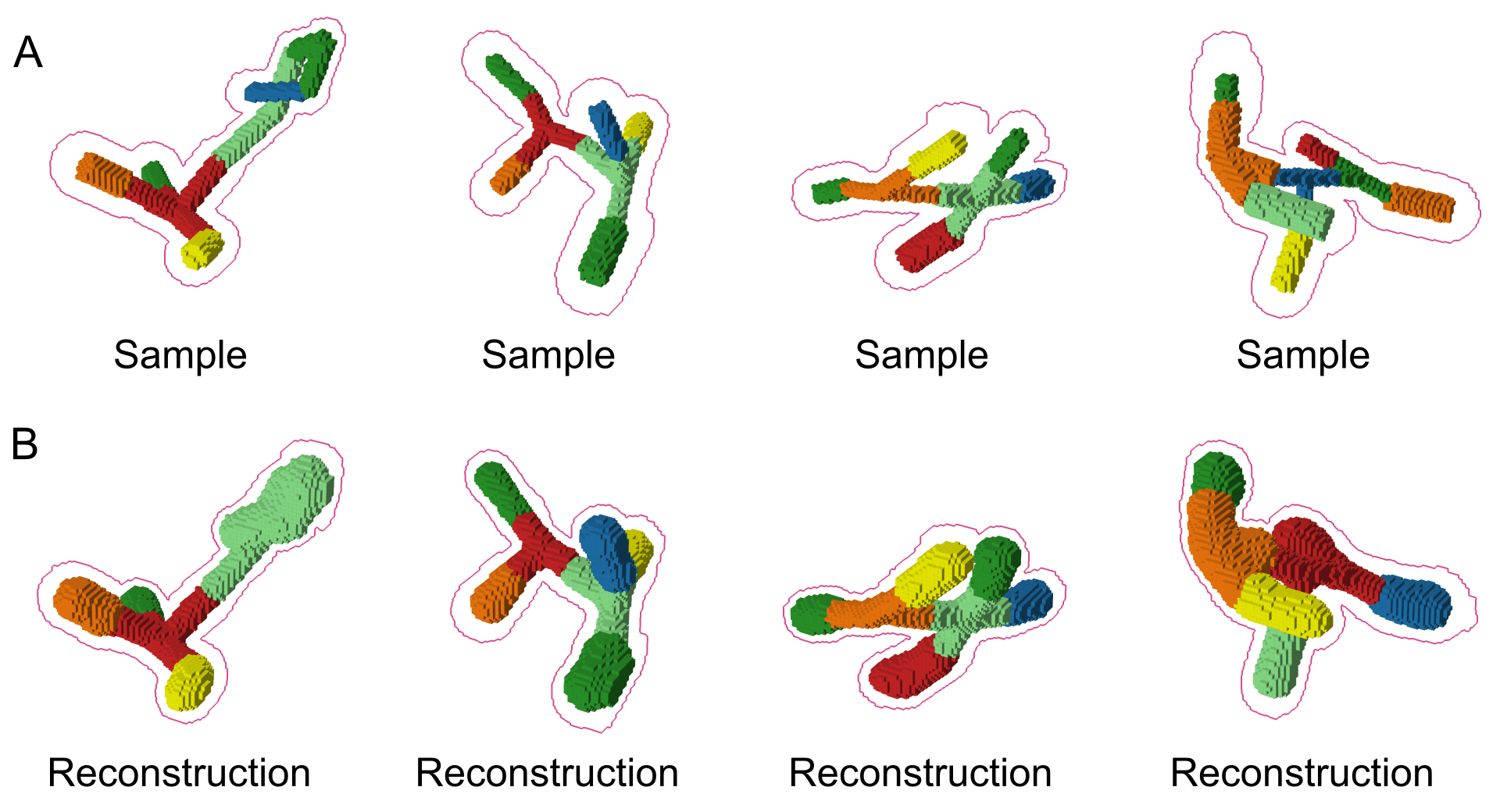}
    \vspace{-16pt}
    \caption{Examples of 
    encoding previously unseen 
    sythentic training data (\textbf{A})
    and 
    reconstructing each one (\textbf{B})
    using the fully optimized VAE.
    Each column displays a sample/reconstruction pair.}
    \label{fig:synth_vs_decode}
\end{figure}

\begin{figure}[ht]
    \centering    \includegraphics[width=\textwidth]{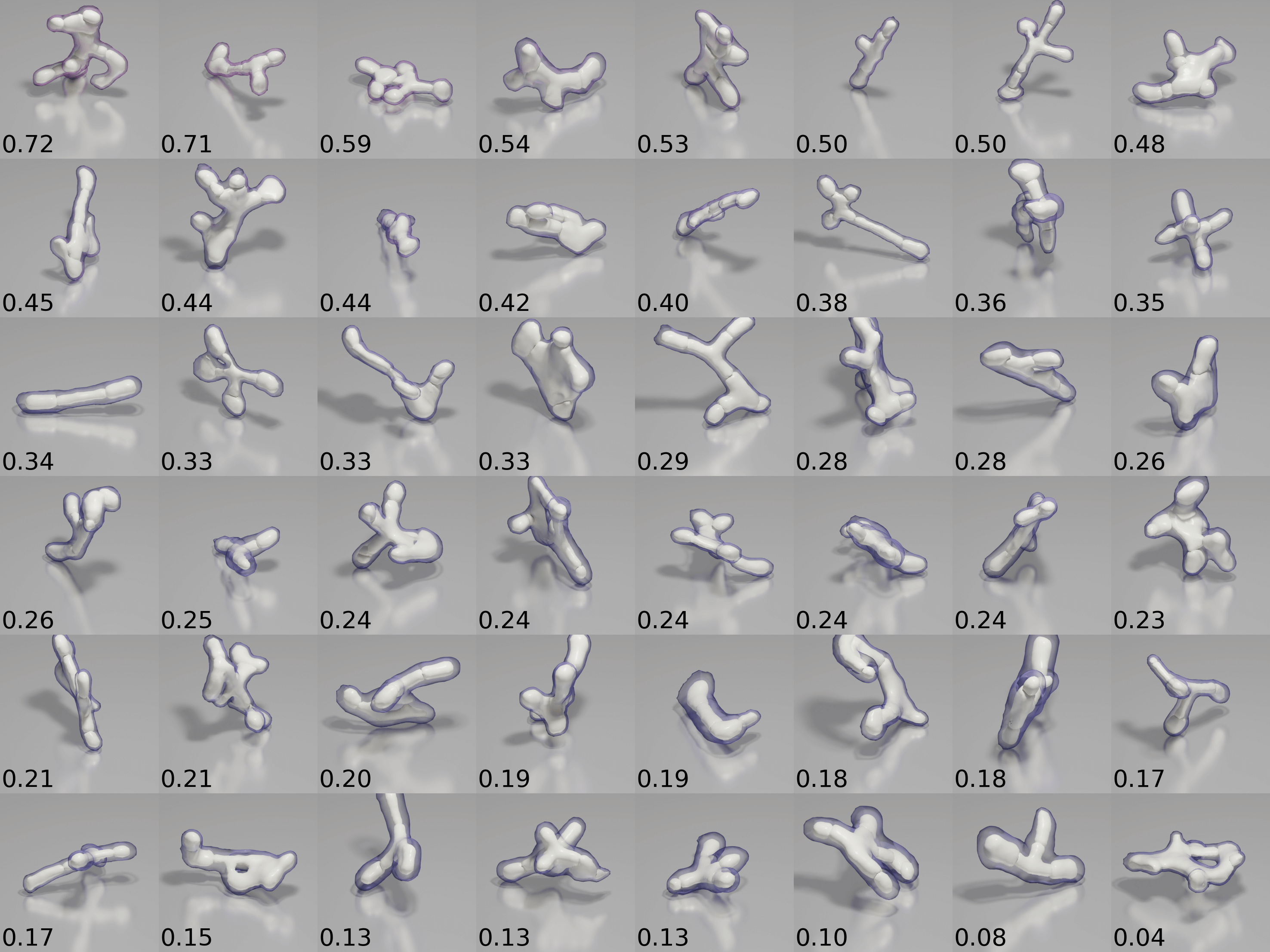}
    \vspace{-18pt}
    \caption{\textbf{Ancestors.}
    Part of a random initial population, sorted by net displacement (in meters).
    }
    \label{fig:standing_grid_init}
\end{figure}

\begin{figure}[ht]
    \centering
    \includegraphics[width=\textwidth]{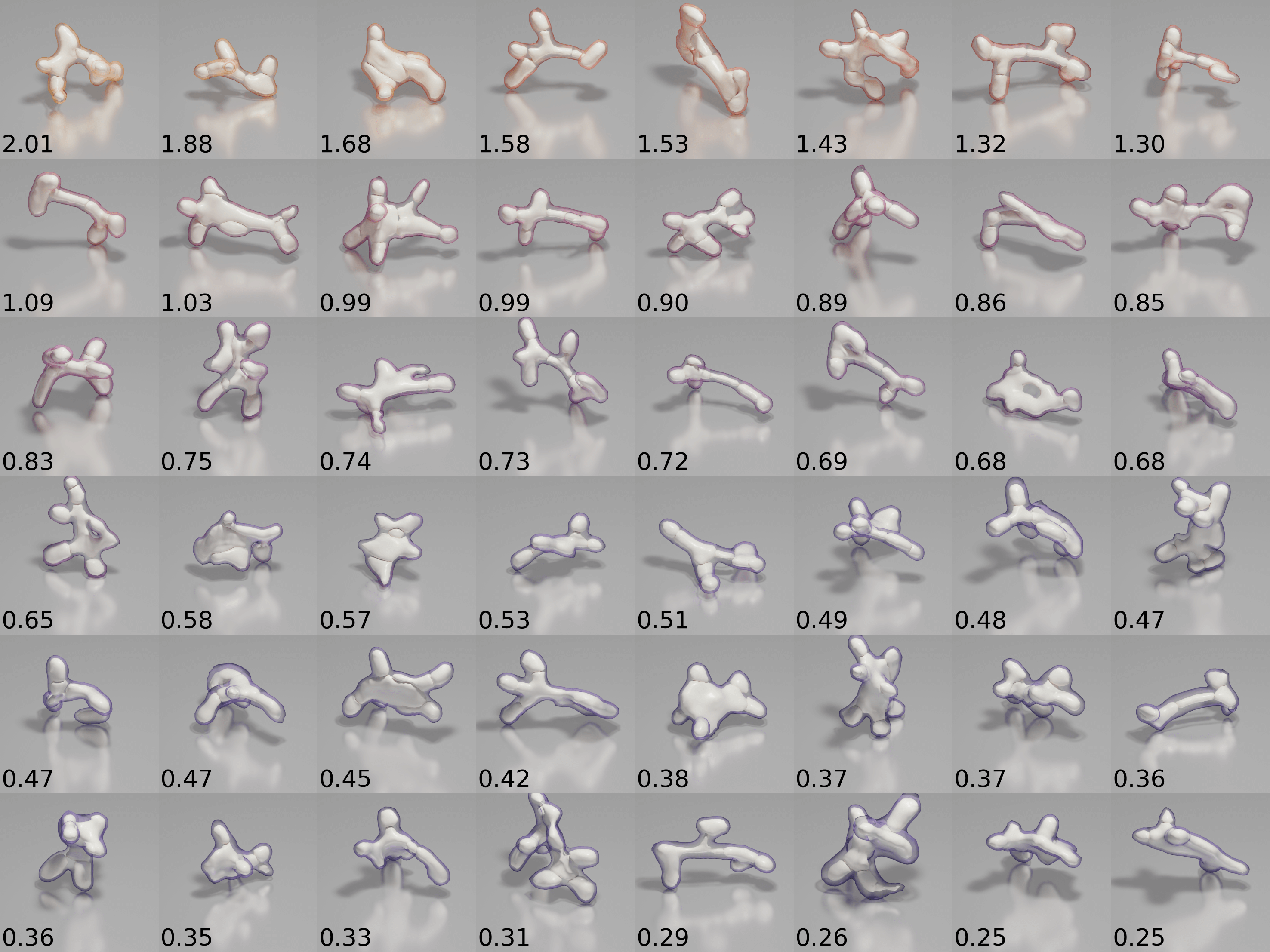}
    \vspace{-18pt}
    
    \caption{\textbf{Descendants.}
    Part of the evolved population that descended from the initial population shown above in Fig.~\ref{fig:standing_grid_init} for Upright Locomotion. Under each design, its net displacement in meters.
    }
    \label{fig:standing_grid_evolved}
\end{figure}

\begin{figure}[t]
    \centering
    \includegraphics[width=\textwidth]{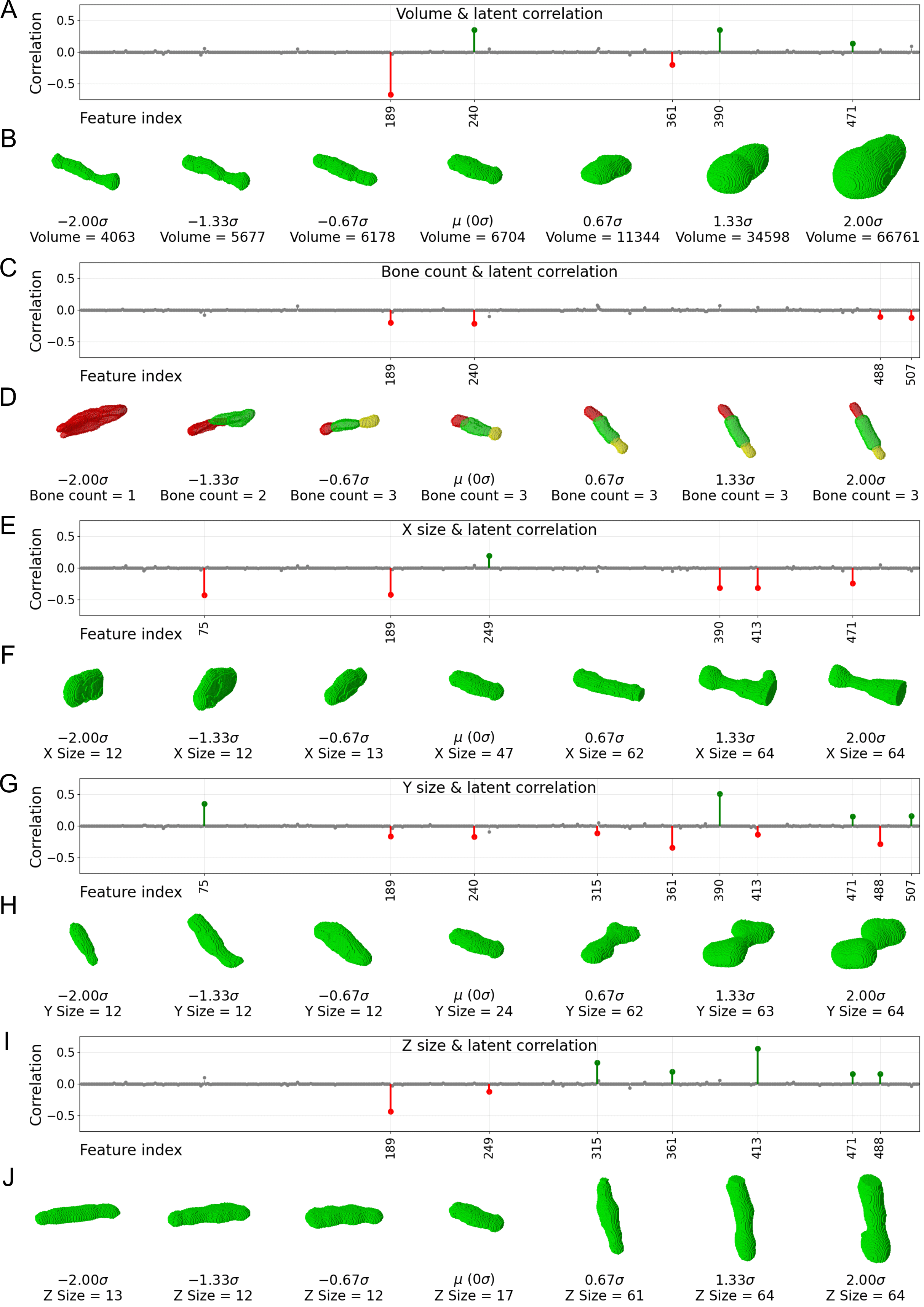}
    \vspace{-16pt}
    \caption{\textbf{Screening for latent genes.} 
    We measured the Pearson correlation coefficient ($\rho$) 
    of the 512 latent features 
    and the
    body volume
    of one million randomly sampled designs (\textbf{A}).
    The set of latent features that 
    are highly correlated with volume ($|\rho|>0.1$)
    were extracted as ``the gene for volume'', the effect of which can be increased or decreased on demand 
    (\textbf{B}).
    The effect of tuning the gene for volume $\pm 2\sigma$ 
    is shown for the
    design decoded from the
    sample mean of the latent space
    ($\bm{\mu}$).
    The same procedure was repeated to identify and modulate 
    genes for 
    bone count (\textbf{C, D}), 
    length (\textbf{E, F}), 
    width (\textbf{G, H}) and 
    height (\textbf{I, J}).
    }
    \label{fig:latent_correlation}
\end{figure}

\begin{figure}[t]
    \centering
    \includegraphics[width=\textwidth]{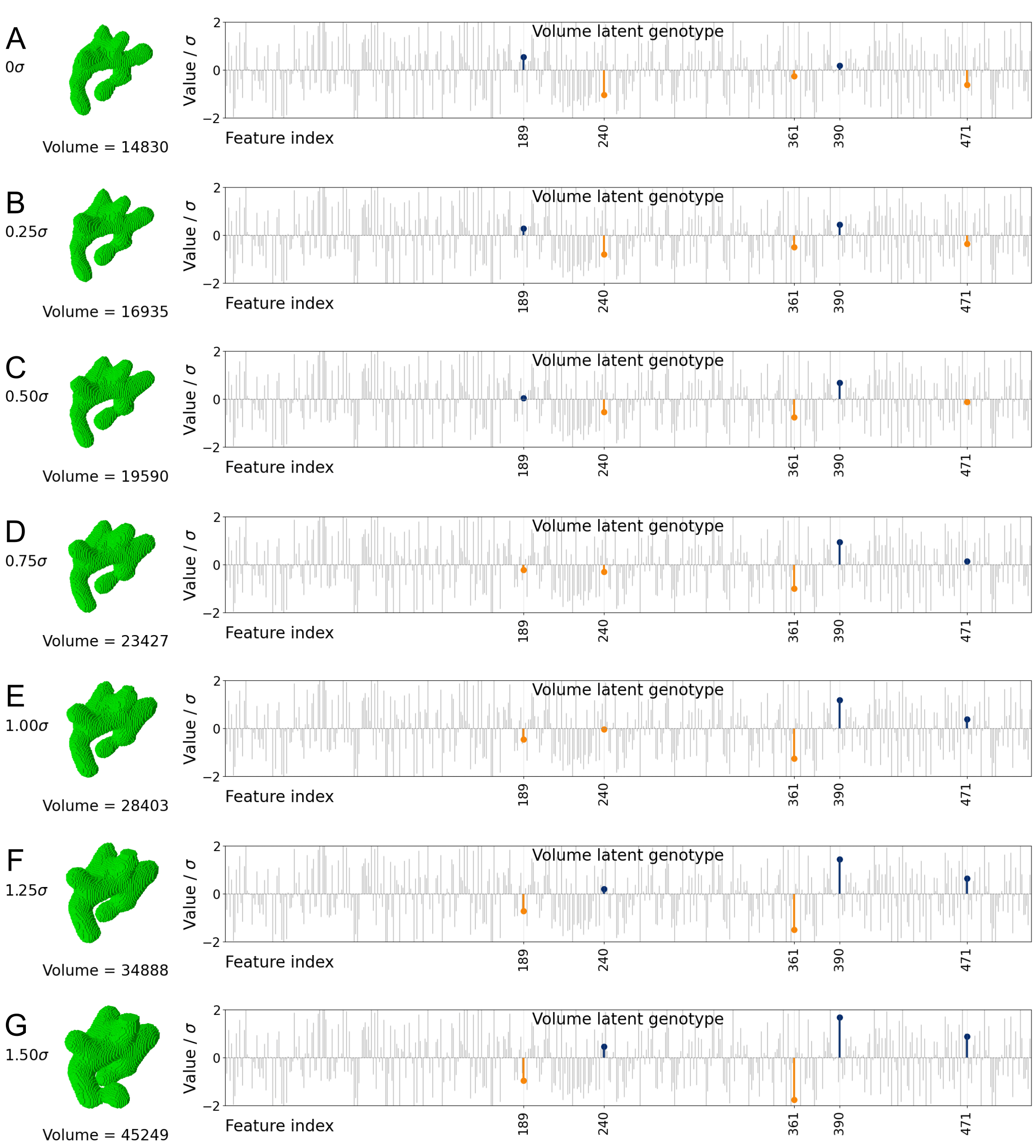}
    \vspace{-16pt}
    \caption{\textbf{Gene editing.} 
    The best design that evolved for Upright Locomotion was extracted (\textbf{A})
    and the set 
    of latent features 
    found to be individually 
    correlated 
    with body volume---the gene for volume---was incrementally increased and the volume of the  resulting design was measured
    (\textbf{B}-\textbf{G}).
    The entire latent genome is shown to the right of each decoded design, 
    with features outside the volume gene grayed out. 
    Positively correlated features (240, 390 and 471)
    were amplified,
    and negatively correlated features (189 and 361) were reversed by
    0.25$\sigma$ in each row, yielding progressively thicker body parts within the evolved body shape.
    }
    \label{fig:vol_gene_editing}
\end{figure}

\begin{figure}[t]
    \centering
    \includegraphics[width=\textwidth]{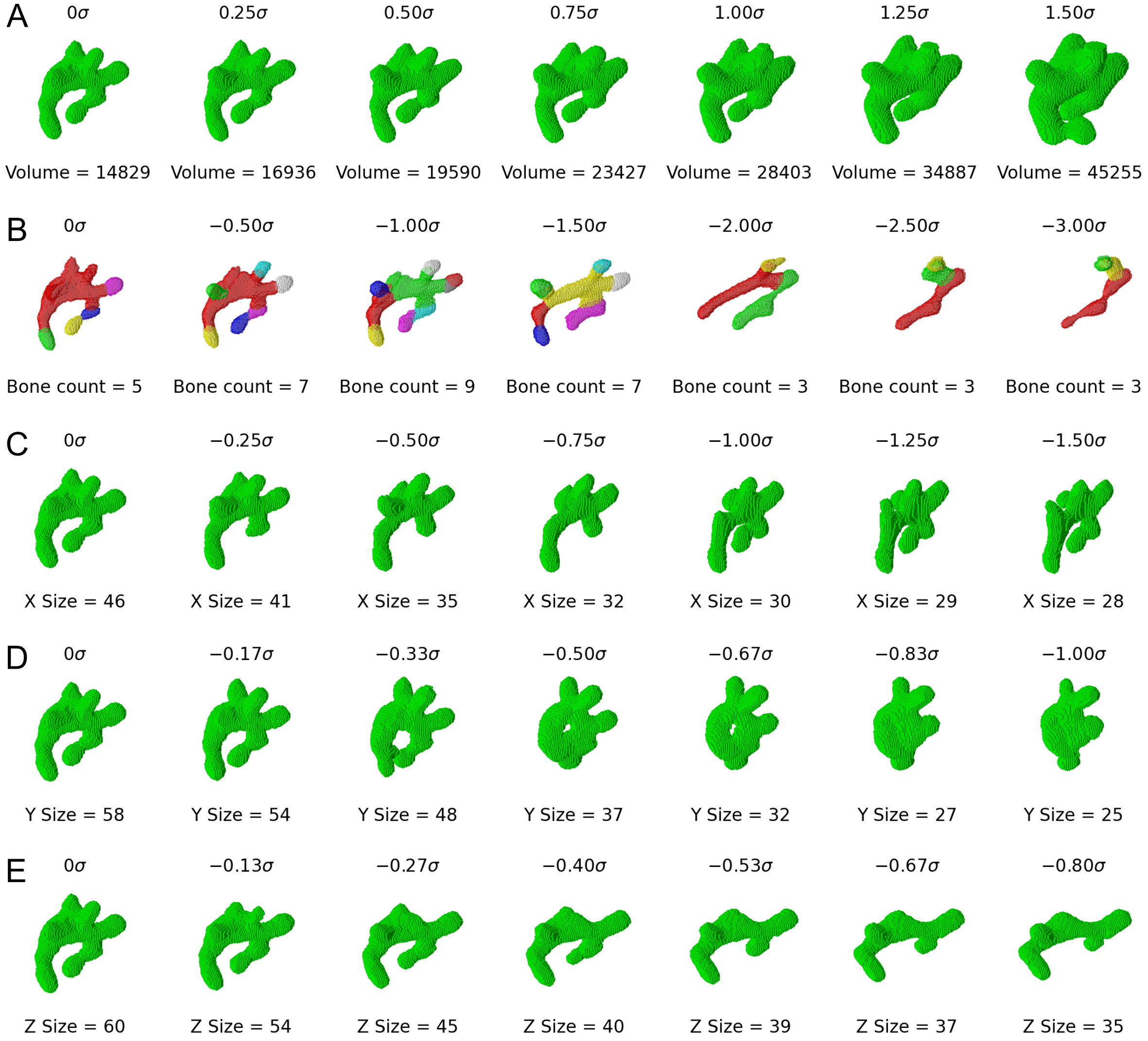}
    \vspace{-16pt}
    \caption{\textbf{Programmable morphology.}
    The best design that evolved for Upright Locomotion was extracted (from Fig.~\ref{fig:env}B,F)
    and the found 
    genes for 
    volume (\textbf{A}),
    bone count (\textbf{B}),
    length (\textbf{C}), 
    width (\textbf{D}) 
    and 
    height (\textbf{E})
    were individually modulated to 
    demonstrate the precision with which evolved morphologies 
    can be manually redesigned 
    through ``genetic engineering''.
    Some genes afford greater control over a specific morphological trait than others; 
    however, this may be attributed to the simplicity of the procedure employed to
    screen for genes, 
    which
    only captures pairwise linear associations between a manually-specified morphological trait
    and an individual latent feature.
    }
    \label{fig:gene_editing}
\end{figure}

\begin{figure}[ht]
    \centering
    \includegraphics[width=0.75\textwidth]{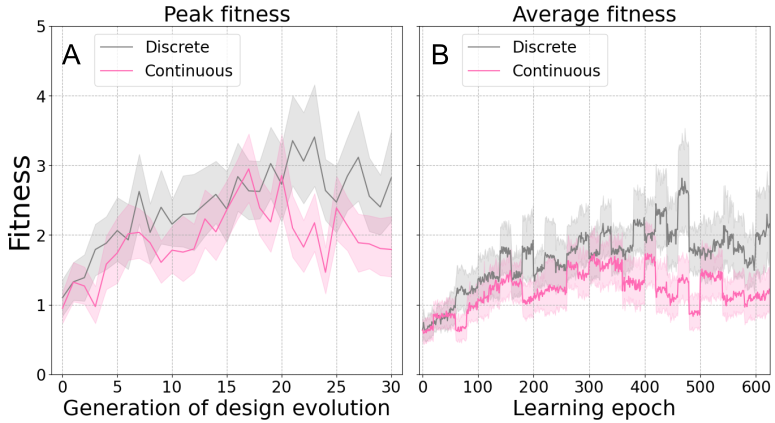}
    \vspace{-4pt}
    \caption{\textbf{Discrete and continuous action spaces.}
    Peak (\textbf{A}) and average (\textbf{B}) fitness is plotted for the Upright Locomotion task, as in Fig.~\ref{fig:before-after-evo}A and B.
    Two kinds of action spaces were compared: discrete (gray) and continuous (pink).
    The discrete action space is \{-1.4 rad, -0.7 rad, 0 rad, 0.7 rad, 1.4 rad\}
    and the continuous action space is [-1.4 rad, 1.4 rad], for all joints.
    In both panels, each line is a mean across the 64 designs in the population and the shaded regions are a 95\% bootstrapped CI of the mean.
    Although both 
    continuous and discrete action spaces 
    are viable,
    a discrete set of actions greatly simplified and stabilized universal policy training
    under the tested conditions.
    }
    \label{fig:discrete_vs_continous}
\end{figure}

\begin{figure}[ht]
    \centering
    \includegraphics[width=0.75\textwidth]{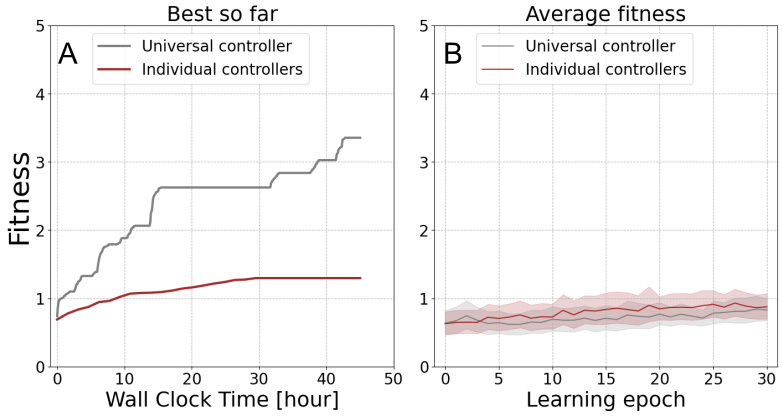}
    \vspace{-4pt}
    \caption{\textbf{Universal and individual controllers.}
    An otherwise equivalent experiment was conducted in which the universal controller shared by all 64 designs in the population
    was replaced with 
    a bespoke individual controller
    for each design.
    As in Fig.~\ref{fig:env},
    the peak fitness achieved 
    by each design was
    averaged across 
    the population
    before plotting the cumulative max (\textbf{A}).
    This statistic, which 
    we term
    ``best so far'',
    captures the progress 
    of the entire population.
    Training 64 independent policies for 30 epochs 
    required 44.8 hours 
    on
    4 NVIDIA H100 SXM GPUs.
    During the same time frame on the same hardware 
    the universal controller can be trained for 451 epochs
    and thereby achieve much higher fitness.
    Over the first 30 epochs of learning, the performance of the universal controller was not statistically different from that of the individual controllers (\textbf{B}), and since design evolution was only advanced a single generation, 
    the designs under universal control were 
    visually similar to those with independent controllers, 
    with both populations 
    closely resembling 
    the random initial population (Fig.~\ref{fig:standing_grid_init}).
    }
    \label{fig:universal_vs_local}
\end{figure}

\clearpage

\section{Simulation}
\label{appx:sim}

In this section we describe the underyling soft and rigid body physics behind our endoskeletal robot simulator.
We will use the symbol $\boldsymbol{\tau}$ for both moment and torque to avoid confusion. Subscripts are used to indicate the source of an effect and superscripts indicate the target of an effect.

\subsection{Soft Voxels} 
Following \cite{hiller2014dynamic},
soft voxels
are defined as a point of mass with rotational inertia in space. 
Each surface of a soft voxel may connect by a Euler–Bernoulli beam with zero mass to an adjacent soft voxel (Fig.~\ref{fig:sim}A) 
or rigid voxel 
(Fig.~\ref{fig:sim}B,C). 
Each beam exerts moment $\boldsymbol{\tau}_b$ and force $\boldsymbol{f}_b$ 
onto the corresponding connected voxel when deformed 
from expansion, compression or twisting.
Although Euler–Bernoulli beams are more complex to compute than Hookean springs, 
the additional torsional moment $\boldsymbol{\tau}_b$ contributes to preserving volume when the structure is under external torsional stress. 
Since endoskeletal robots embed joints between rigid bones inside soft tissue and produce large deformations, 
preserving volume prevents generating invalid force and moment values 
when soft tissues are radically compressed 
and thereby stabilizes simulation.

The moment $\boldsymbol{\tau}_b$ and force $\boldsymbol{f}_b$ generated by beams can be computed from position $\mathbf{x} = (x, y, z)$ and rotation $\boldsymbol{\theta}$ of voxels $v_1$ and $v_2$ attached on both ends using a constant stiffness matrix $K$:
\begin{equation}
\begin{bmatrix}\label{eq:beam}
\boldsymbol{f}_b^{v1} \\
\boldsymbol{\tau}_b^{v1} \\
\boldsymbol{f}_b^{v2} \\
\boldsymbol{\tau}_b^{v2}
\end{bmatrix}
=
[K]
\begin{bmatrix}
\mathbf{x}^{v1} \\
\boldsymbol{\theta}^{v1} \\
\mathbf{x}^{v2} \\
\boldsymbol{\theta}^{v2}
\end{bmatrix}
\end{equation}
Following Newton's second law, we can express the equation of motion for soft voxels as:
\begin{equation}
\begin{aligned}\label{eq:soft_newton}
m^v \ddot{\mathbf{x}}^v &= \sum \boldsymbol{f}_b^v + \boldsymbol{f}_{\text{ext}}^v \\
I^v \ddot{\boldsymbol{\theta}}^v &= \sum \boldsymbol{\tau}_b^v + \boldsymbol{\tau}_{\text{ext}}^v
\end{aligned}
\end{equation}
where $f_{\text{ext}}^v$ and $\tau_{\text{ext}}^v$ represent external force and moment caused by gravity, friction, and contacting force etc. For simplicity, we will define generalized mass, position and force representations $M^v$, $X^v$ and $F^v$ and rewrite  Eq.~\ref{eq:soft_newton} as:
\begin{equation}\label{eq:generalized_soft_newton}
M^v \ddot{X}^v = \sum F_b^v + F_{\text{ext}}^v \quad \text{with} \quad X^v = \begin{bmatrix} \mathbf{x}^v \\ \boldsymbol{\theta}^v \end{bmatrix}, F^v = \begin{bmatrix} \boldsymbol{f}^v \\ \boldsymbol{\tau}^v \end{bmatrix}, M^v = \begin{bmatrix} m^v E & 0 \\ 0 & I^v \end{bmatrix}
\end{equation}
where $E$ is the identity matrix.

\subsection{Rigid Voxels}

Rigid voxels are affected by the beams of connected soft voxels (Fig.~\ref{fig:sim}C), 
generalized external forces $F_{\text{ext}}^v$, 
and additional generalized constraint forces $F_c^v$, as from joints (Fig.~\ref{fig:sim}D).
Since rigid voxels belonging to the same rigid bone $r$ do not have relative motion, 
we compute the generalized position $X^r$ of the compound rigid bone $r$ and infer $X^v$. 
As each rigid bone will be connected by many beams to its surrounding soft voxels, 
we will use $F_b^r$ as the total generalized force exerted by all beams to simplify the representation:
\begin{equation}\label{eq:generalized_rigid_newton}
M^r \ddot{X}^r = F_b^r + F_{\text{ext}}^r + F_c^r
\end{equation}

For $F_b^r$ and $F_{\text{ext}}^r$, 
torque is also generated by forces applied to the center of component rigid voxels.
Thus:
\begin{equation}\label{eq:rigid_detail}
F_b^r = \sum_{v \in r, b \in v} \begin{bmatrix} \boldsymbol{f}_b^v \\ \boldsymbol{\tau}_b^v + (\mathbf{x}^v - \mathbf{x}^r) \times \boldsymbol{f}_b^v \end{bmatrix}
\quad \text{and} \quad
F_{\text{ext}}^r = \sum_{v \in r} \begin{bmatrix} \boldsymbol{f}_{\text{ext}}^v \\ \boldsymbol{\tau}_{\text{ext}}^v + (\mathbf{x}^v - \mathbf{x}^r) \times \boldsymbol{f}_{\text{ext}}^v \end{bmatrix}
\end{equation}
where
$x_r$ represents the center of mass of the corresponding rigid bone $r$. 

Contacts were modeled between voxel pairs
and contact forces
were thus aligned to the center of colliding voxels instead of the center of mass of the rigid bones.
To prevent generating 
a false torque term in $F_{\text{ext}}^v$, 
the contact force does not contribute to the torque.

Finally, constrained dynamics were used to compute
$F_c^r$ as follows.
Constraints were defined as functions between two rigid bones $r_1$ and $r_2$:
\begin{equation}\label{eq:constraint}
c(X^{r_1}, X^{r_2}) \geq 0
\end{equation}
For convenience, we now remove superscripts 
and use concatenated $X$, $F$, and block-diagonalized $M$ to represent attributes of both bones. 
By taking the time derivative of constraint function $c$ we obtain the Jacobian matrix $J$ which maps global space to constraint space:
\begin{equation}\label{eq:constraint_jacobian}
\dot{c}(X) = J \dot{X}
\end{equation}
Introducing Lagrange multiplier $\lambda$, which physically represents applied constraint forces, and we can rewrite Eq.~\ref{eq:generalized_rigid_newton} as:
\begin{equation}\label{eq:constraint_lambda}
M \ddot{X} = F + J^T \lambda \quad \text{with} \quad F =  F_b + F_{\text{ext}}
\end{equation}

By substituting Eq.~\ref{eq:constraint_lambda} into Eq.~\ref{eq:constraint_jacobian} we get:
\begin{equation}\label{eq:constraint_target}
\dot{c}(X) = J\dot{X_{\text{old}}} + JM^{-1}(F + J^T\lambda)\Delta t
\end{equation}
where $\dot{X_{\text{old}}}$ is speed from the last time step; this derivative needs to be non-negative if we wish the constraint to be satisfied at last time point. 
We also add a bias factor 
\citep{baumgarte1972stabilization}
to correct constraint errors incrementally:
\begin{equation}\label{eq:constraint_target_final}
\dot{c}(X) = J\dot{X_{\text{old}}} + JM^{-1}(F + J^T\lambda)\Delta t + \frac{\beta}{\Delta t}c_{\text{old}}(X) \geq 0
\end{equation}
where $\beta$ is a constant that controls the constraint error correction rate in each time step. 

Reordering terms in Eq.~\ref{eq:constraint_target_final} and we obtain equation system in the form of a linear complementarity problem (LCP) and then solve $\lambda$ using the projected Gauss-Seidel (PGS) algorithm \citep{erleben2013numerical}:
\begin{eqnarray}
A &=& JM^{-1}J^T, \\
w &=& J\dot{X_{\text{old}}} + JM^{-1}F\Delta t + \frac{\beta}{\Delta t}c_{\text{old}}(X), \\
A\lambda + w &\geq& 0 \\
\lambda &\geq& 0 \label{eq:lambda_non_neg_condition}\\
\lambda &\perp& A\lambda + w \label{eq:lambda_perp_condition}
\end{eqnarray}

Note that Eq.~\ref{eq:lambda_non_neg_condition} is required since $\lambda$ is in constraint space 
and it should be in the positive direction that satisfies the constraint. 
Note also that Eq.~\ref{eq:lambda_perp_condition} is required since constraint forces does not add energy to the system and thus should always be perpendicular to the direction of the constraint change.


\section{From Simulation to Reality: Future Work}
\label{appx:sim2real}



The endoskeletal robots in this paper were confined to simulated tasks within a virtual world.
Though significant effort was dedicated to ensuring
that the
simulated
mechanics,
simulated
environments
and
simulated
materials
closely resembled physical reality
and realizable morphologies (Table~\ref{table:sim_hypers}),
it is in general difficult to guarantee that the behavior of a system optimized in simulation will transfer with sufficient fidelity to reality \citep{jakobi1995noise}.
It remains to determine the effect that universal controllers have on sim2real transfer, since sim2real of universal control has yet to be attempted.
But because our approach simultaneously optimizes a population of many robots with differing topological, geometrical, sensory and motor layouts, 
it realizes the desired behavior in many unique ways 
and thereby provides many more opportunities for successful sim2real transfer
compared to other methods that 
optimize a single design \citep{kriegman2020scalable}.
We also predict that the inherent capacity of a universal controller to generalize across these differing 
action and observation spaces 
will
render the policy 
more transferable across 
certain differences between 
simulation and reality,
compared to a bespoke single-morphology controller.

\clearpage

\section{Overview of the Entire Co-design Pipeline}
\begin{figure}[ht]
    \centering
    \includegraphics[width=\textwidth]{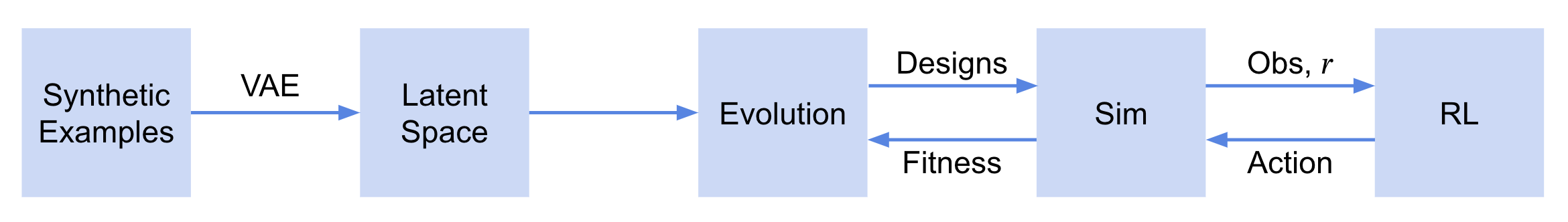}
    \vspace{-18pt}
    \caption{\textbf{Data flow.} Synthetic examples are used to encode the latent search space in which evolution operates.
    An evolving population of designs decoded from this space
    are simulated and
    used to train a controller.
    The population is updated 
    based on the fitness 
    of the designs.
    }
    \label{fig:overview}
\end{figure}

\section{Synthetic Data Generation}
\label{appx:synth_data_gen}

\begin{algorithm}[h]
    \caption{Procedural Generation of Synthetic Robot Body Plan}
    \begin{algorithmic}[1]
        \State \textbf{Input:} Node range $[K_{\text{min}}, K_{\text{max}}]$, degrade ratio $\psi$
        \State Randomly select number of nodes $k$ in $[K_{\text{min}}, K_{\text{max}}]$
        \For{each node $i = 1$ to $k$}
            \State Randomly initialize:
            \State \quad Max children $C_i$
            \State \quad Limb length $L_i$
            \State \quad Rigid radius $r_{\text{rigid}, i}$
            \State \quad Soft radius $r_{\text{soft}, i}$
            \State Create \textit{StarNode} $i$
        \EndFor
        \State Set node $1$ as root (level $0$)
        \For{each node $i = 2$ to $k$}
            \State Select parent node $p$ with lowest level that can attach more children
            \State Decrease limb length: $L_i = L_i \times \psi^{\text{level}_p}$
            \State Compute offset vector opposite to $p$'s existing children, blend with random unit vector
            \State From $p$, use $L_p$ and offset to find attachment point; use $L_i$ to place center of node $i$
            \State Attach node $i$ to parent node $p$
        \EndFor
    \end{algorithmic}
\end{algorithm}

\begin{table}[h]
\centering

\caption{Hyperparameters and their ranges used in the procedural generation of synthetic robot body plans.}

\vspace{4pt}

\begin{tabular}{|l|l|p{8cm}|}
\hline
\textbf{Parameter} & \textbf{Range / Values} & \textbf{Description} \\
\hline
\multicolumn{3}{|l|}{\textit{Global Parameters}} \\
\hline
$K_{\text{min}}$ & 3 & Minimum number of nodes (stars) per robot. \\
$K_{\text{max}}$ & 8 & Maximum number of nodes (stars) per robot. \\
$\psi$ & $0.9$ & Degradation ratio multiplier for limb length per hierarchical level. \\
\hline
\multicolumn{3}{|l|}{\textit{Node Parameters (for each node $i$)}} \\
\hline
$C_i$ & \{2, 3, 4\} & Maximum number of children (arms) for node $i$. \\
$L_i$ & $[0.1, 0.3]$ & Limb length for node $i$. \\
$r_{\text{rigid}, i}$ & $[0.02, 0.05]$ & Radius of the rigid (bone) parts for node $i$. \\
$r_{\text{soft}, i}$ & $[0.02, 0.07]$ & Radius of the soft (tissue) parts for node $i$. \\
\hline
\end{tabular}
\end{table}

\clearpage

\section{More Hyperparameters}
\label{appx:more_hypers}


\begin{table}[ht]
\centering

\caption{VAE \textit{encoder} architecture hyperparameters.}

\vspace{4pt}

\begin{tabular}{|c|l|}
\hline
\textbf{Layer}  & \textbf{Description} \\ \hline
Input           & Voxel input, \( 64^3 \times (k + 2) \) \\ \hline
VRN0            & Voxception-ResNet (VRN), kernel size \( 3^3\), 32 channels \\
Block 1         & 5x VRN, 32 channels, downsample to \( 32^3 \), 64 channels \\
Block 2         & 5x VRN, 64 channels, downsample to \( 16^3 \), 128 channels \\
Block 3         & 5x VRN, 128 channels, downsample to \( 8^3 \), 256 channels \\
Block 4         & 5x VRN, 256 channels, downsample to \( 4^3 \), 512 channels \\
ResConv         & Residual Conv, kernel size \( 3^3\), 512 channels \\
Max Pool     & Max pooling, 512 channels\\
Linear             & in: 512, out: 1024, GeLU \\
Linear-mu, Linear-logvar & mu: in: 512, out: 512, logvar: in: 512, out: 512\\ \hline
Output & mu: 512, logvar: 512 \\ \hline
\end{tabular}
\label{tab:vae_encoder}
\end{table}

\begin{table}[ht]
\centering

\caption{VAE \textit{decoder} architecture hyperparameters.}

\vspace{4pt}

\begin{tabular}{|c|l|}
\hline
\textbf{Layer}  & \textbf{Description} \\ \hline
Input           & Embedding input, 512 \\ \hline
Linear           & in: 512, out: 32768, reshaped to \( 4^3 \times 512 \) \\ 
Block 1         & 5x VRN, 512 channels, ConvTranspose3D to \( 8^3 \), 256 channels \\
Block 2         & 5x VRN, 256 channels, ConvTranspose3D to \( 16^3 \), 128 channels \\
Block 3         & 5x VRN, 128 channels, ConvTranspose3D to \( 32^3 \), 64 channels \\
Block 4         & 5x VRN, 64 channels, ConvTranspose3D to \( 64^3 \), $k$ + 2  channels (voxel prediction)\\ \hline
Output & Voxel logits, \( 64^3 \times (k + 2) \) \\
\hline
\end{tabular}

\label{tab:vae_decoder}
\end{table}

\begin{table}[h]
\centering
\caption{RL encoder architecture hyperparameters. 
The RL encoder is shared by the actor and critic, combining spatial pooling and graph transformer.}
\vspace{4pt}
\begin{tabular}{|c|l|}
\hline
\textbf{Layer} & \textbf{Description} \\ \hline
Input & $s^v$ (4 channels), $s^r$ (14 channels), $s^j$ (9 channels)\\ \hline
ResNet (voxels wise) & in: 4, out: 64 \\ 
\makecell{Spatial Encoder-Decoder Net \\ 
\cite{peng2020convolutional}} & \makecell{ Feature grid dim: $64^3 \times 64$ \\ UNet3D levels: 2 \\ UNet3D groups: 8 \\ UNet3D feature maps: 64} \\ 
TransformerConv 1 & node\_in: 78 (64 + 14), node\_out: 256, edge\_in: 9 \\ 
TransformerConv 2 & node\_in: 256, node\_out: 512, edge\_in: 9 \\ 
TransformerConv 3 & node\_in: 512, node\_out: 512, edge\_in: 9 \\ \hline
Output & \makecell{$\tilde{\mathbf{n}}_t$: 512 \\ $\tilde{\mathbf{e}}_t$ : 1024 (by concatenate node features $\tilde{\mathbf{n}}_t$)} \\ \hline
\end{tabular}
\label{tab:rl_encoder}
\end{table}

\begin{table}[ht]
\centering
\caption{Actor architecture hyperparameters.}
\vspace{4pt}
\begin{tabular}{|c|l|}
\hline
\textbf{Layer} & \textbf{Description} \\ \hline
Input & $\tilde{\mathbf{e}}_t$ Outputs from RL encoder (see Table~\ref{tab:rl_encoder}) \\ \hline
ResNet 1 & in: 1024, out: 256 \\ 
ResNet 2 & in: 256, out: 64 \\ 
ResNet 3 & in: 64, out: $||\mathcal{A}||$ \\ \hline
Output & $\mathbf{z}_t$: $||\mathcal{A}||$ \\ \hline
\end{tabular}
\label{tab:rl_actor}
\end{table}

\begin{table}[ht]
\centering
\caption{Critic architecture hyperparameters.}
\vspace{4pt}
\begin{tabular}{|c|l|}
\hline
\textbf{Layer} & \textbf{Description} \\ \hline
Input & $\tilde{\mathbf{n}}_t$ Outputs from RL encoder (see Table~\ref{tab:rl_encoder}) \\ \hline
Max Pool & Max pooling, 512 channels \\ 
ResNet 1 & in: 512, out: 128 \\ 
ResNet 2 & in: 128, out: 32 \\ 
ResNet 3 & in: 32, out: 1 \\ \hline
Output & $V(s_t)$: 1 \\ \hline
\end{tabular}
\label{tab:rl_critic}
\end{table}

\begin{table}[ht]
\centering

\caption{Simulation configuration hyperparameters.}
\label{table:sim_hypers}
\vspace{4pt}

\begin{tabular}{|c|c|}
\hline
\textbf{Parameter}            & \textbf{Value}  \\
\hline
\multicolumn{2}{|l|}{\textit{Environment}} \\
\hline
Rigid solver iterations       & 10             \\ 
Baumgarte ratio ($\beta$)     & 0.01           \\ 
Gravity acceleration          & 9.81 m/s$^2$         \\ 
Control signal frequency      & 10 Hz           \\ 
Max joint motor torque        & 6 N*m           \\
Voxel size                    & 0.01 m           \\ 
\hline
\multicolumn{2}{|l|}{\textit{Soft Material}} \\
\hline
Elastic modulus               & $3 \times 10^4$ Pa  \\ 
Density                       & 800 kg/m$^3$     \\ 
Poisson's ratio               & 0.35           \\ 
Static friction               & 1            \\
Dynamic friction              & 0.8            \\
\hline
\multicolumn{2}{|l|}{\textit{Rigid Material}} \\
\hline
Density                       & 1500 kg/m$^3$    \\ 
Static friction               & 1            \\ 
Dynamic friction              & 0.8            \\ \hline
\end{tabular}

\end{table}

\begin{table}[ht]
\centering

\caption{Reinforcement learning and evolution hyperparameters.}

\vspace{4pt}

\begin{tabular}{|c|c|}
\hline
\textbf{Parameter}            & \textbf{Value}  \\
\hline
\multicolumn{2}{|l|}{\textit{RL Hyperparameters}} \\
\hline
Actor learning rate   & $6 \times 10^{-5}$  \\ 
Critic learning rate & $6 \times 10^{-5}$  \\ 
Discount ($\gamma$) & 0.9 \\
Entropy weight & 0.01 \\
GAE & Not used \\
Learning epochs & 20 \\ 
PPO train batch size & 64  \\ 
PPO train steps per epoch & 60  \\
$P_{\text{small}}$ & 0.1 [voxels] \\
$P_{\text{large}}$ & 10 [voxels] \\
Minimum movement threshold ($\delta$) & 0.001 m\\ 
\hline
\multicolumn{2}{|l|}{\textit{Population and Rollouts}} \\
\hline
Population size ($P_{\text{large}}$)  & 64  \\ 
Clones per individual ($\text{rollout\_num}$) & 2 \\
\hline
\end{tabular}
\end{table}



\begin{table}[t]
\centering
\caption{Symbols.}

\label{table:symbol_table}

\vspace{4pt}

\begin{tabular}{|l|l|}
\hline
\textbf{Symbol} & \textbf{Meaning} \\ 
\hline
\multicolumn{2}{|l|}{\textit{Simulation}} \\
\hline
$m$ & Mass \\ 
$x, y, z$ & Position of each axis in space \\
$\epsilon$ & Average strain of a soft voxel \\ 
$\mathbf{x}$ & Position in space \\ 
$\mathbf{v}$ & Linear velocity \\ 
$\boldsymbol{\omega}$ & Angular velocity of a rigid bone \\ 
$\mathbf{q}$ & Quaternion representing the orientation\\
$\boldsymbol{\theta}$ & Rotation in space \\ 
$\boldsymbol{\tau}$ & Moment or torque \\ 
$\boldsymbol{f}$ & Force \\ 
$X$ & Generalized position\\ 
$\dot{X}$ & Generalized velocity \\ 
$\ddot{X}$ & Generalized acceleration \\ 
$\dot{X}_{\text{old}}$ & Generalized velocity from the previous time step \\
$M$ & Generalized mass matrix \\ 
$K$ & Stiffness matrix of a beam \\ 
$I$ & Inertia \\  
$F$ & Generalized force vector \\ 
$c(X)$ & Constraint function between two rigid bones \\ 
$J$ & Jacobian matrix \\ 
$\lambda$ & Lagrange multiplier \\ 
$A$ & Matrix in the LCP \\ 
$w$ & Term involving velocities and external forces in the LCP \\ 
$\Delta t$ & Time step size in simulation\\ 
$\beta$ & Constraint error correction rate \\ 
\hline
\multicolumn{2}{|l|}{\textit{Reinforcement Learning}} \\
\hline
$r$ & Reward\\
$s$ & State\\ 
$a$ & Action\\
$V$ & Value function\\ 
$\pi$ & Policy function\\ 
$\gamma$ & Discount factor \\ 
$\mathbf{z}$ & Action logits\\ 
$\mathcal{A}$ & Action space \\ 
$\mathbf{n}$ & Node features \\
$\mathbf{e}$ & Edge features \\
$\theta$ & Joint angle \\ 
$\mathbf{h}$ & Hinge axis direction of a joint \\ 
$\mathbf{d}_1, \mathbf{d}_2$ & Vectors from joint to the center of mass of connected bones \\ 
$\mathbf{u}$ & History movement vector \\
$\mathbf{v}$ & Movement vector between two time steps, or linear velocity \\ 
\hline 
\multicolumn{2}{|l|}{\textit{Utilities}} \\
\hline
$\mathbb{I}$ & Indicator function \\
$E$ & Identity matrix \\ 
\hline
\multicolumn{2}{|l|}{\textit{Subscript and Superscript Conventions}} \\
\hline
$b$ & Beam (subscript) \\ 
ext & External (subscript) \\ 
$t$ & Time step in RL (subscript) \\ 
$v$ & Soft voxel (superscript) \\ 
$r$ & Rigid bone (superscript) \\ 
$j$ & Joint (subscript/superscript) \\
$l$ & Local pool (superscript) \\
\hline
\end{tabular}
\end{table}

\end{document}